\documentclass[letterpaper, 10 pt, journal, twoside]{IEEEtran}
\usepackage[utf8]{inputenc}
\usepackage{amssymb,graphicx,epstopdf,amsmath,color,algorithm,tabularx,float,url,gensymb}
\usepackage{placeins}
\usepackage{hyperref,dblfloatfix,verbatim}
\usepackage{subcaption,cleveref,overpic}
\usepackage[T1]{fontenc} 
\usepackage{array,multirow,graphicx}
\usepackage{float}
\usepackage{ctable}
\usepackage{colortbl}
\usepackage{varwidth}
\usepackage{algorithm}
\usepackage{algpseudocode}
\usepackage[export]{adjustbox}
\usepackage[super,comma]{natbib}

\newcommand{\STAB}[1]{\begin{tabular}{@{}c@{}}#1\end{tabular}}
\newcolumntype{?}{!{\vrule width .12em}}
\newcommand{\rom}[1]{\uppercase\expandafter{\romannumeral #1\relax}}
\DeclareMathOperator*{\argmax}{arg\,max}

\title{Tactile Model O: Fabrication and testing of a 3D-printed, three-fingered tactile robot hand}
\author{Jasper W. James*, Alex Church*, Luke Cramphorn and Nathan F. Lepora%
\thanks{The authors are with the Dept. of Engineering Mathematics,
        University of Bristol, Bristol, UK and The Bristol Robotics Laboratory, Bristol, UK.
        {\tt\small \{jj16883, ac14293, n.lepora\}@bristol.ac.uk}}%
\thanks{* The two lead authors contributed equally to the research in this paper.}
}

\begin{document}

\maketitle
\begin{abstract}
Bringing tactile sensation to robotic hands will allow for more effective grasping, along with the wide range of benefits of human-like touch. Here we present a 3D-printed, three-fingered tactile robot hand comprising an OpenHand Model~O customized to house a TacTip soft biomimetic tactile sensor in the distal phalanx of each finger. We expect that combining the grasping capabilities of this underactuated hand with sophisticated tactile sensing will result in an effective platform for robot hand research -- the Tactile Model O (T-MO). The design uses three JeVois machine vision systems, each comprising a miniature camera in the tactile fingertip with a processing module in the base of the hand. To evaluate the capabilities of the T-MO, we benchmark its grasping performance using the Gripper Assessment Benchmark on the YCB object set. Tactile sensing capabilities are evaluated by performing tactile object classification on 26 objects and predicting whether a grasp will successfully lift each object. Results are consistent with the state of the art, taking advantage of advances in deep learning applied to tactile image outputs. Overall, this work demonstrates that the T-MO is an effective platform for robot hand research and we expect it to open-up a range of applications in autonomous object handling. Supplemental video: \url{https://youtu.be/RTcCpgffCrQ}.
\end{abstract}

\section{Introduction}
Tactile afferents in our hands provide information about the state of a grasp and, crucially, whether the grasp is failing.\citep{johansson2009coding} Therefore, bringing tactile sensation to robotic hands will allow for more effective grasping, along with the wide range of benefits of human-like touch. This will enable tasks we take for granted yet robots are presently incapable of achieving.

Tactile enabled robots will be able to operate autonomously and safely in cluttered and unknown environments in such varied situations as health and social care or industrial manufacture.\citep{Kappassov2015TactileReview} \textcolor{black}{Tactile sensing has been demonstrated to be useful in multiple areas of robotics including single-point sensors such as whiskers, large area sensors and high resolution fingertips.\citep{luo2017robotic}} \textcolor{black}{However significant scientific and social barriers remain that prevent tactile robot hands from becoming commonplace. These include the development of robust and safe soft sensors, and public acceptance of robots for use in domestic settings.\citep{Lee2000TactileChallenges}} 

Although robotic hands with integrated soft tactile sensors are becoming more common,\citep{Yousef2011,Kappassov2015TactileReview} the vast majority of research on robotic hands still focuses on vision-guided grasp planning. In part, this is because there are a limited selection of commercial tactile-enabled hands, which are expensive and behind the development curve of the most advanced tactile sensors. Thankfully, the advent of fast, precise multimaterial 3D-printing technologies now means it is relatively straightforward and inexpensive to adapt tactile sensors and robotic hands into new integrated platforms with both soft and hard parts. However, scaling to multiple sensors on a robotic hand presents many challenges, such as how to control the hand with integrated sensor feedback. With those challenges in mind, the presented platform offers a balance of dexterous capability against relative simplicity in its underactuation and fabrication.

The aim of this study is to modify a three-fingered robotic hand, the OpenHand Model O,\citep{odhner2014compliant} to house the TacTip, a soft optical tactile sensor with 3D-printed biomimetic morphology, \citep{ward2018tactip} such that the combined system capable of predicting whether a grasp will be stable and identify objects using tactile sensing alone. Prior work on integrating this tactile sensor with 3D-printed hands has involved two 2-DoF, two-fingered robotic hands: the OpenHand model-M2 gripper with a tactile thumb and the GR2 gripper with two tactile fingertips.\citep{ward2016tactile, ward2017gr2} The current work represents a major advance in the functionality of 3D-printed tactile dexterous hands to integrate three tactile fingertips within a 4-DoF hand with state-of-the-art grasping capabilities. Here we demonstrate tactile capabilities by performing object classification and predicting the stability of grasps using CNNs applied to tactile data alone. We expect that the grasping capabilities of the Model O combined with the success of the TacTip as a tactile sensor will result in an effective platform -- the tactile Model~O (T-MO) -- for a wide range of future tactile robot hand applications and research.

\textcolor{black}{The main contributions of this study are to:
\begin{enumerate}
    \item Customize the design of a 3D-printed tactile sensor (the TacTip) for integration with a 3D-printed robot hand designed for grasp research (the Model O).
    \item Assess the performance of this tactile hand on real-time one-shot grasp success prediction and object classification, using deep learning on tactile images.
    \item Assess also the sensitivity of this performance to a reduced number of tactile sensors, to demonstrate the utility of tactile sensing within a robot hand.
\end{enumerate}}
\noindent\textcolor{black}{An area of further novelty is that the new tactile sensor design contains an on-board processing unit that enables local computation. For example, data processing for each finger can be done on the hand, in principle encompassing deep learning models for perception (via TensorFlow Lite).} Overall, the T-MO's combined capability at grasping and tactile perception results in a low-cost tactile grasping system which sets a new level of capability in the field of robot hands.
\begin{figure*}[t]
\centering
    	\begin{overpic}[height=8cm,trim={0cm 0cm 0cm 0cm},clip=true]{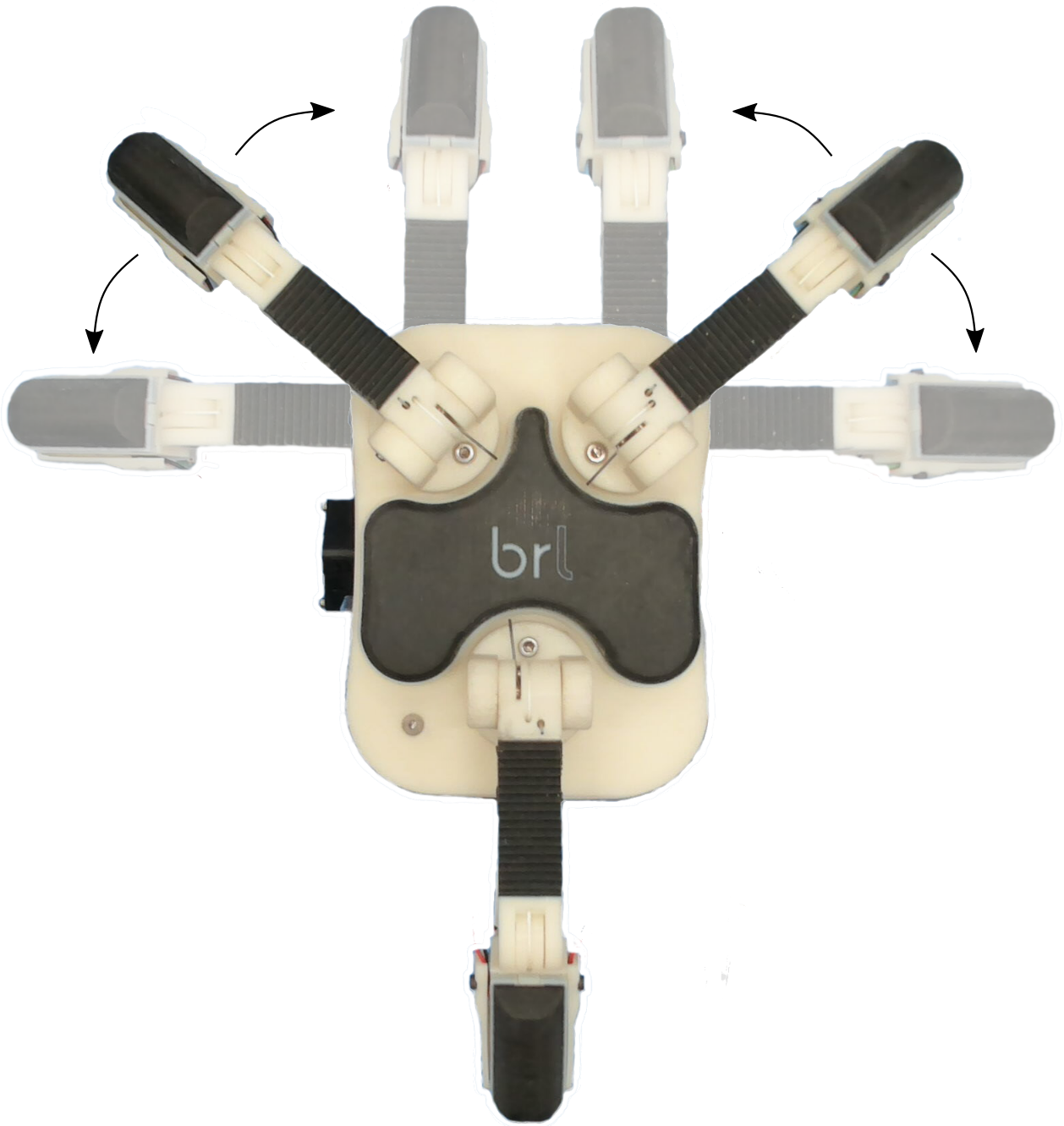}
        \put(34,7){(1)}
        \put(10,91){(2)}
        \put(80,91){(3)}
        \end{overpic}
\includegraphics[height=8cm]{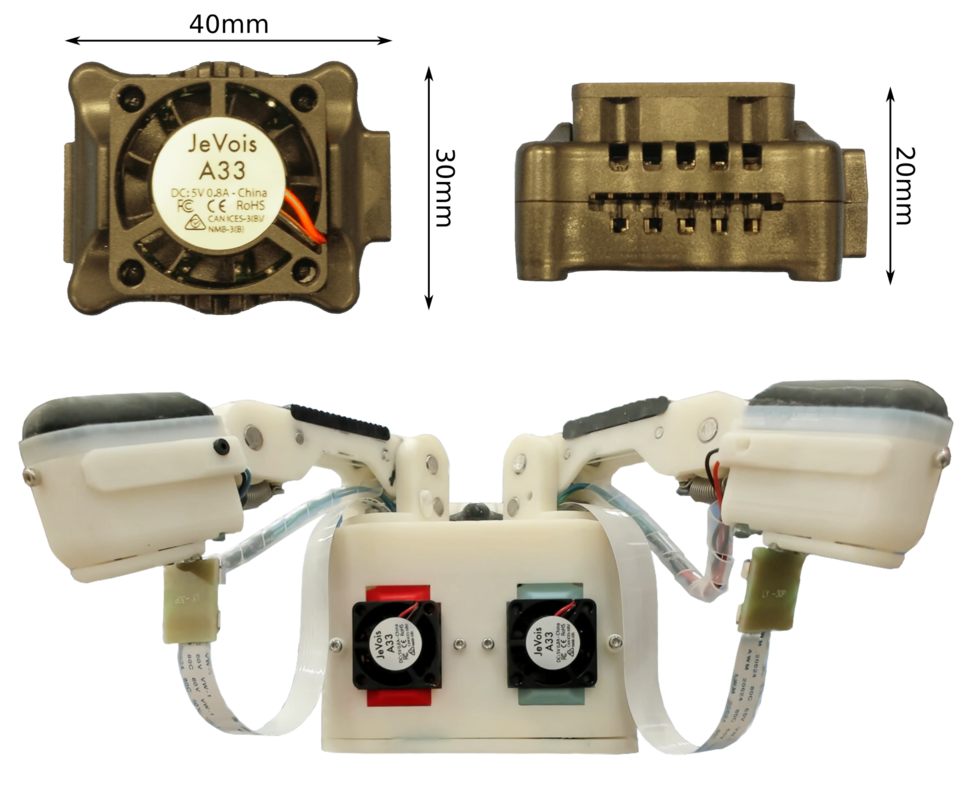}
    \caption{Left: The Tactile Model O (T-MO) with fingers labelled (1-3). Note the range of movement of fingers (2,3), which are mechanically coupled giving a single degree of freedom. Right: JeVois machine vision modules above a side view of the T-MO. Four Dynamixel servos and three JeVois are housed in the base unit with a marginal increase in size compared to the  Model~O. The lower image shows how the JeVois processors are integrated into the base of the hand. A ribbon cable runs from each JeVois unit to its camera module in the distal phalanx of each finger. A second cable supplies electricity to the LEDs.}
        \label{fig:top_modelo_sf}
\end{figure*}

\section{Background}
The majority of work with tactile robot hands has involved solid-state tactile sensors, and has been reviewed extensively elsewhere.\citep{Kappassov2015TactileReview,Yousef2011} Widely-used platforms include the anthropomorphic iCub hand with integrated capacitive sensors,\citep{Schmitz2010AICub} the Shadow hand with BioTac sensors placed at the fingertips \citep{Abd2018DirectionHand} and with fabric and capacitive tactile sensors,\citep{Koiva2013ASensor} the 4-fingered Allegro hand with PPS capacitive sensors \citep{Jara2014ControlFeedback} and with BioTac sensors~\citep{veiga2018hand} and the TWENDY-ONE hand \citep{Iwata2009DesignTWENDY-ONE} with embedded force-torque sensors and a tactile skin; \citep{Schmitz2014TactileDropout} this last hand was the first to use deep learning for object recognition. In addition, the i-HY hand used in this study has been integrated with MEMS barometric TakkTile sensors, to give a basic array of 4 pressure-sensitive taxels on each fingertip. \citep{odhner2014compliant}

Several studies have integrated optical tactile sensors onto two-fingered robotic grippers. The first was \citet{ward2016tactile}, who used the M2 gripper, a two-fingered two-DoF hand from the OpenHand project similar to the hand considered here~\citep{Ma2016M2Gripper}, and rolled cylinders across the surface using only tactile feedback. Next, two GelSight optical tactile sensors were integrated onto a two finger parallel gripper and used for slip detection\citep{dongimproved}. Subsequently, \citet{ward2017gr2} integrated the TacTip onto a GR2 two-fingered gripper, \citep{rojas2016gr2} also rolling cylinders along a trajectory in the hand's workspace over the TacTip's surface. Then followed a slimmer version of the GelSight, the GelSlim, which reflects light down the finger to a camera module at its base and has a flat profile appropriate for parallel jaw grippers\citep{Donlon2018GelSlim:Finger}. There recently followed a more compact, two-fingered robot gripper with multiple Gelsight optical tactile sensors covering the inner surface of the hand \citep{wilson2020design}.

\textcolor{black}{In relation to the present study with TacTip optical tactile sensors integrated into a 3-fingered hand, both the M2 and GR2 grippers are effective at establishing pinch grasps but their design is focused on the in-hand manipulation of objects through simple control which makes them unsuitable for many of the tasks performed here.}



Testing of tactile enabled hands has largely taken the form of collecting a dataset of grasps on various objects and using various machine learning methods to try to distinguish between them. \citet{Spiers2016Single-GraspSensors} use 11 objects from the YCB object set using data from an array of barometric pressure sensors attached to a two-fingered hand and classify them using random forests. They obtained a validation accuracy of $94\%$ when the objects' orientations were unconstrained. \citet{Flintoff2018Single-GraspSensors} also use random forests on a two-fingered hand containing barometric sensors and the Google Soli radar sensor to classify $26$ objects, obtaining $99\%$ validation accuracy. \textcolor{black}{In this work the hand was placed on a surface with objects placed in the sample plane with minor variation in object positions and orientations.} \citet{Schmitz2014TactileDropout} identify a set of $20$ objects with an accuracy of $88\%$ using a deep neural network on the four-fingered TWENDY-ONE hand. \textcolor{black}{Significantly, it was shown that dropout can be used with tactile data to notably improve performance.}

\textcolor{black}{\citet{regoli2017controlled} achieve an accuracy of 97.6\% using a kernel regularised least-squares method on data collected over 21 objects from the YCB set. Here objects are placed into an iCub hand and tactile data from two fingers is gathered during exploratory grasping and wrapping actions. A grasp stabilisation phase resulted in an increase of up to 29\% accuracy on object recognition. \citet{FunabashiObjectRecognition} use uSkin tactile sensors on an Allegro hand alongside proprioceptive joint angles as the input to a CNN. Objects are placed into the hand and 95\% accuracy is obtained when classifying 20 objects: 10 from the YCB set and 10 cylindrical objects. \citet{LiuObjectRecognition} investigate whether the intrinsic relationship between tactile data collected from different fingers can be exploited to improve object recognition performance. Two object recognition tasks are used, classifying 5 bottles using a BarrettHand with capacitive tactile sensors; and classifying up to 10 household objects using a Schunk parallel gripper and piezoresistive tactile sensors. The best accuracy over these tasks was found using the a joint kernel sparse coding method that exploits the similarity between data gathered on individual fingers via a shared sparsity support pattern.}

\textcolor{black}{Another common task for evaluating tactile robot hands is grasp stability prediction. Data-driven approaches are often used in the form of collecting a dataset of successful and unsuccessful grasps, then using this information to train machine learning algorithms. \citet{WanVariabilityGrasping} predicted grasp stability on an iHY robotic hand with MEMS barometers but use only one object at a fixed position, making grasps highly repeatable, and obtained an accuracy of 90\% using an SVM. Similarly, \citet{dang2014stable} use a BarrettHand with capacitive sensors to train an SVM. Simulated data from 704 objects resulted in 81\% accuracy. A physical experiment was also performed on 6 household objects, where grasps were sampled until the simulation trained SVM predicts a successful grasp, at which point an attempt to raise to object is performed. Using this method a grasp success rate of 84.6\% was achieved, although the baseline success rate in the training data was not given. \citet{krug2016analytic} applied an analytic approach by using tactile data to determine whether a set of wrenches (concatenated force/moment vectors) necessary to grasp an object can be applied by the hand. This approach required no training data and was computationally efficient. When applied to a test dataset of 584 grasps distributed over 4 household objects the classification accuracy was 74\%.}

\textcolor{black}{Most prior work on object recognition and grasp success prediction has used either objects passed directly to a fixed robotic grasper, or a fixed grasper constrained to the same plane as the object. Often there is variation added to the object position and orientation to make the task more challenging, although this is typically at the discretion of the user. There are a limited number of studies that perform object classification in a more realistic pick-and-place environments, such as the one considered here. We are of the view that such environments are more representative of practical applications.}

\section{Methods}
\subsection{Tactile Sensor}
This study involves the integration of a soft optical biomimetic tactile sensor in the TacTip family \citep{ward2018tactip,chorley2009development} onto a OpenHand Model O, an underactuated three fingered hand.\citep{odhner2014compliant} The TacTip surface consists of a rubber-like 3D-printed surface made from Tango Black+ and supported at its base by a hollow cylinder made from Verro White plastic. The sensor tip is filled with silicone (RTV27905) and sealed with a clear acrylic lens. This design allows the sensor to deform around objects and regain its shape post-contact. 

The principal design aspect of the TacTip family is an array of protruding pins arranged inside the sensor surface, which mimic the dermal papillae and Merkel cell complexes present in the glabrous (non-hairy) skin of primate fingertips. The pins are printed on the inside of the hemisphere using Tango Black+ and a small tip of Vero White is printed on top of the Tango Black+ `rods' to make pin detection easier to visualize from a camera. The entire TacTip exterior is printed as a single unit, using both materials, which has allowed complicated pin structures to be tested that would be very difficult to manufacture with other fabrication methods.\citep{Cramphorn2017AdditionAcuity} 

White LEDs are used to give uniform illumination inside the sensor and reduce the effects of any external light that bleeds through the Tango Black+ skin or Vero White base. A camera is mounted on the base of the sensor to view the pins as the sensor deforms and either the pin positions or the raw camera images can be used as the tactile information, for example as tactile inputs into supervised learning methods for regression or classification over labelled data.

\begin{figure}[t]
\centering
\includegraphics[width=0.45\textwidth, trim={0cm 0cm 0cm 0cm}, clip=true]{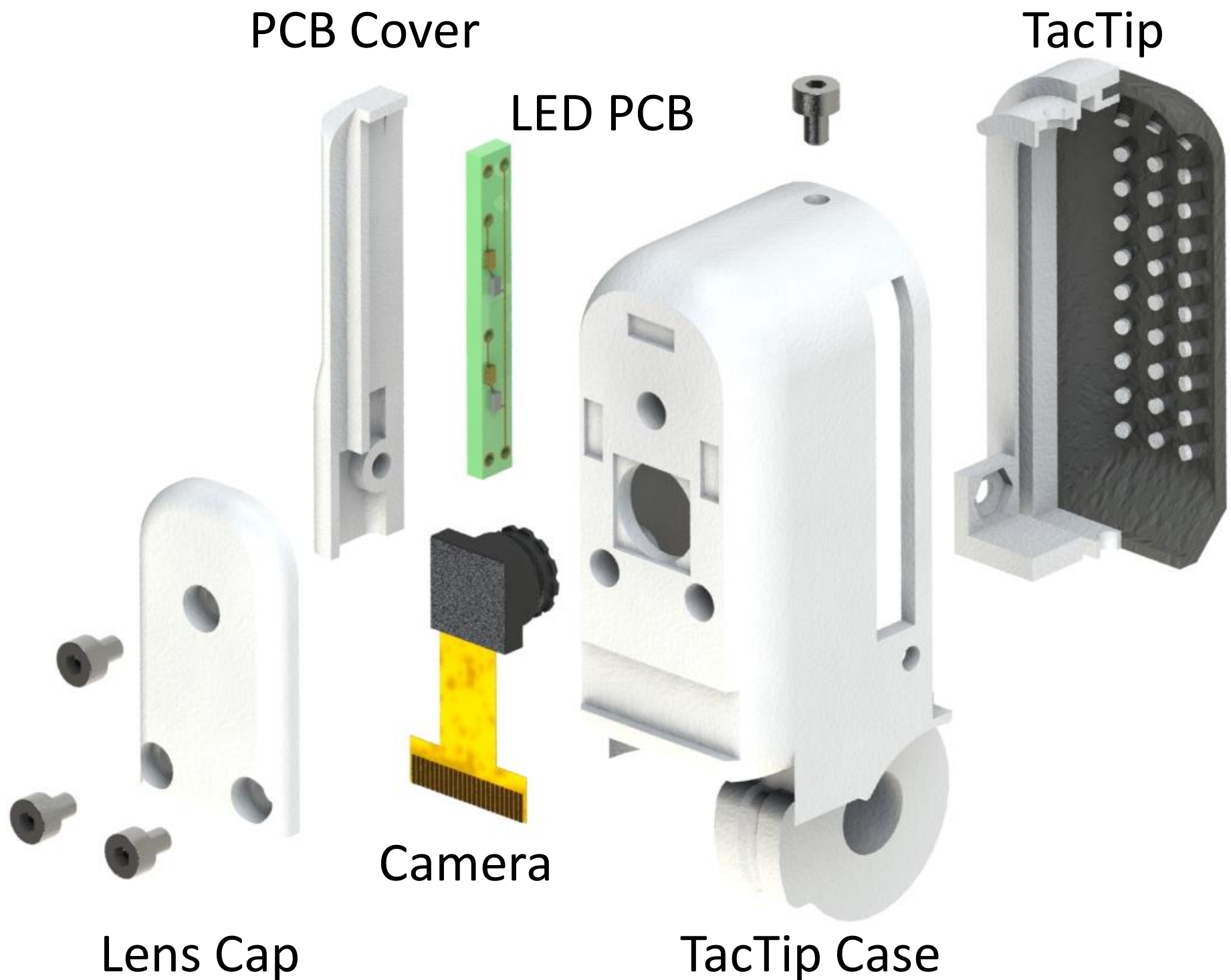}
\caption{Exploded view of the distal phalanx in the T-MO finger.}
\label{fig:explode}
\end{figure}
\subsection{Robotic Hand}
Underactuated robot hands greatly simplify the control systems required to grasp objects by using their morphology to conform to objects.\citep{Catalano2014AdaptiveSoftHand,Ajoudani2014ExploringSoftHand} The hand chosen for this work was the Model O, developed by the OpenHand project at Yale.\citep{ma2017yale} The Model~O is based on the i-HY hand first presented by \citet{odhner2014compliant} which won the ARM-H track of the DARPA manipulation challenge.\citep{Hackett2013AnProgram}. The Model~O is a three-fingered under-actuated hand with four DoF which, along with other OpenHand designs, has demonstrated great success in grasping by leveraging the morphology of its design.\citep{odhner2014compliant,Ma2013AHand} The Model O is mostly 3D-printed (using ABS plastic) and completely open source. This makes it ideal for this study as manufacture and modification of the design for integration of the tactile sensor is straightforward.

The three fingers are almost identical in that they contain two joints and a single degree of freedom, making each underactuated in the same way. A braided polyethylene wire `tendon' runs the length of the finger and is connected to a Dynamixel MX-28T motor in the base of the hand. This allows the fingers to deform around objects for grasping without the object shape being known by the controller. Springs in both joints cause the finger to passively release when the active force from the motor is removed.

The only difference between the fingers is that one, the `thumb', is fixed to the palm of the hand, whereas the other two fingers can rotate through 90 degrees from facing the thumb to facing each other (Fig.~\ref{fig:top_modelo_sf}). The rotation of these two fingers is mechanically coupled and therefore constitutes only a single degree of freedom.

\subsection{Modification of Finger Design for Sensor Integration}
We had several goals for the design when modifying the Model O to integrate the TacTip tactile sensor:
\begin{enumerate}
\item To maintain the strong grasping performance of the Model O by keeping the tendon driven actuation design.
\item To minimise the size of sensor such that the Model O distal phalanx was minimally enlarged and strong sensing performance was retained.
\item To integrate the camera processing unit into the motor housing of the Model O with minimal size increase.
\textcolor{black}{\item To keep the system low cost and easily fabricated via 3D printing.}
\end{enumerate}
\noindent With these design goals in mind, we now discuss the changes we have introduced to the Model O hand.

Each finger of the Model O has two joints and two phalanges with pads constructed from Vytaflex 30 to give a high friction surface suitable for grasping. We have replaced the distal phalanx of each finger with a tactile sensor such that three sensors are used in total (Fig.~\ref{fig:top_modelo_sf}). As the tendon is fixed at the base of the distal phalanx and runs down the centre of the finger, modifying only the distal phalanx allows us to keep the fundamental actuation principals of the Model O, in-keeping with design goal 1. Also, this keeps the complexity of the modification low but still involves collecting data from more TacTips than before. \textcolor{black}{To date, this is the smallest TacTip variant that has been integrated onto a robotic hand.}

Previously, the TacTip had a hemispherical shape with 127 pins. \citep{ward2017gr2,cramphorn2017addition,lepora2017exploratory} However, for integration onto the Model O the sensor was modified to be closer to the shape of the distal finger phalanx of the Model O (Fig.~\ref{fig:explode}). \textcolor{black}{We chose to change the shape to rectangular rather than keeping a smaller hemispherical design. This was to minimise changes to the morphology of the distal phalanx finger pad, with which a hemispherical sensor would necessarily have a smaller sensing area if integrated within a rectangular linkage. A rectangular sensor covers as much of the potential contact area as possible.} 

This reduction in size and change of shape (40\,mm diameter hemisphere to $40\times20$\,mm rectangle) presents two challenges. The first challenge is that the pin layout needs modifying to be consistent with the shape of the distal phalanx. The number of pins in the sensor is reduced to 30, arranged in three rows of 10, to ensure good coverage of the interior surface whilst also having sufficient separation of the pin tips to be easily distinguishable. Four LEDs are mounted above the sensor lens in two strips along the major axis of the finger (previously, six were used in a ring). The pins have a diameter of $1.2$\,mm with centres $3$\,mm apart.

The second challenge involves integrating the camera system: the small size of the distal phalanx means that for the camera to be attached directly above the sensor, the form factor of the camera board must be small and the lens must have both a wide field-of-view and a small focal length. These requirements led to us using a $90\degree$ non-distortion lens connected to a JeVois machine vision camera system.\citep{Itti2019JeVoisCamera} The camera module is connected to the finger (Fig.~\ref{fig:explode}) and a ribbon cable connects it to the JeVois board which is housed in the `base' of the hand where the motors are housed~(Fig.~\ref{fig:top_modelo_sf}). Fundamentally, the new sensor has similar design principles as the original hemispherical TacTip, with a camera observing changes in a soft surface as it contacts objects. While we anticipate some changes in sensing capabilities between the original and modified designs that could be examined in future work, the key focus here is to examine the performance of this new design situated in the hand.

The JeVois can capture frames and process them using OpenCV at a variety of frame rates and resolutions ranging from $1280\times 1024$ (15 FPS) to $176\times 144$ (120 FPS). TensorFlow Lite models can even be loaded directly onto the JeVois, enabling processing by pre-trained deep networks within the hand itself. We use a single JeVois module for each of the three TacTips. We were able to integrate the three Jevois without changing the frontal area of the base by removing the taper between the `palm' of the Model O and the bottom of the base. \textcolor{black}{The JeVois is low cost (\textasciitilde £50) and combining this with the small size means that it satisfies design goals 2-4.} For this study, the base of the hand has a connector unit, for mounting on a six degree-of-freedom robotic arm (UR5, Universal Robots), described in Section \ref{platform} on the Autonomous Grasping Platform, below.

The camera is held in place by screwing a cap over the top of the board. The ribbon cable emerges from the rear of this cap and feeds directly into the base of the hand where the JeVois processors are housed. The TacTip slots into the phalanx and is held by three screws. This makes replacing the sensor skin very simple when breakages inevitably happen. Overall, the integration of the tactile sensors makes the fingertips heavier, the effect of which is compensated by using stronger springs to hold the fingers in a fully relaxed position when the hand is held with the palm facing down.

Whilst the new fingers are roughly the same width and length as those of the original Model O, they are deeper because the use of a camera with a $90\degree$ field-of-view means that the lens must be at least 20\,mm away from the TacTip to view the entire 40\,mm surface. This makes the fingertips significantly thicker than those of the model O, increasing from 12\,mm to 35\,mm. This has the effect of removing the ability to slide the fingertips under objects to initiate grasps. For this reason, we expect that the larger size of the fingers presented here will have an impact on the ability of the hand to pick up very small flat objects, but otherwise the functionality of the hand should be similar to the original Model O.

\subsection{Tactile Data}\label{tactiledata}
\textcolor{black}{There are two modes of operation available for processing tactile data on the T-MO. The first is to track the movement of pins via the Python OpenCV function `SimpleBlobDetector' using a script on the JeVois and pass a list of positions to the control PC using a serial connection. This mode uses the pin positions as the tactile output,\citep{ward2018tactip,lepora2015superresolution} as in previous work integrating the TacTip into 2-fingered grippers.\citep{ward2016tactile,ward2017gr2} When using this mode the JeVois was capable of performing pin detection at the same rate as image capture (60 fps at 320x240px) which provides a major benefit to the system as performing pin detection on three simultaneous video streams is a resource heavy task for a PC. Other off-the-shelf image processing boards such as the Raspberry Pi 3 or Zero were unsuitable as they were too large (Pi 3) or unable to detect pins at a high frame rate (Pi Zero).}

\textcolor{black}{The second technique is to capture the raw camera image and feed this into a neural network with minimal pre-processing. For this mode the control PC can detect each JeVois as a video source and capture frames directly.} Recent research has shown that this second technique gives improved results in a contour following task, particularly when concerned with robustness in an online setting.\citep{Lepora2019FromSensor} Thus, in the present paper, the main approach will be to use raw images and neural networks; the hand has, however, been designed to accommodate both approaches.
\begin{figure}[t]
\centering
\includegraphics[width=0.49\textwidth, trim={0cm 0cm 0cm 0cm}, clip=true]{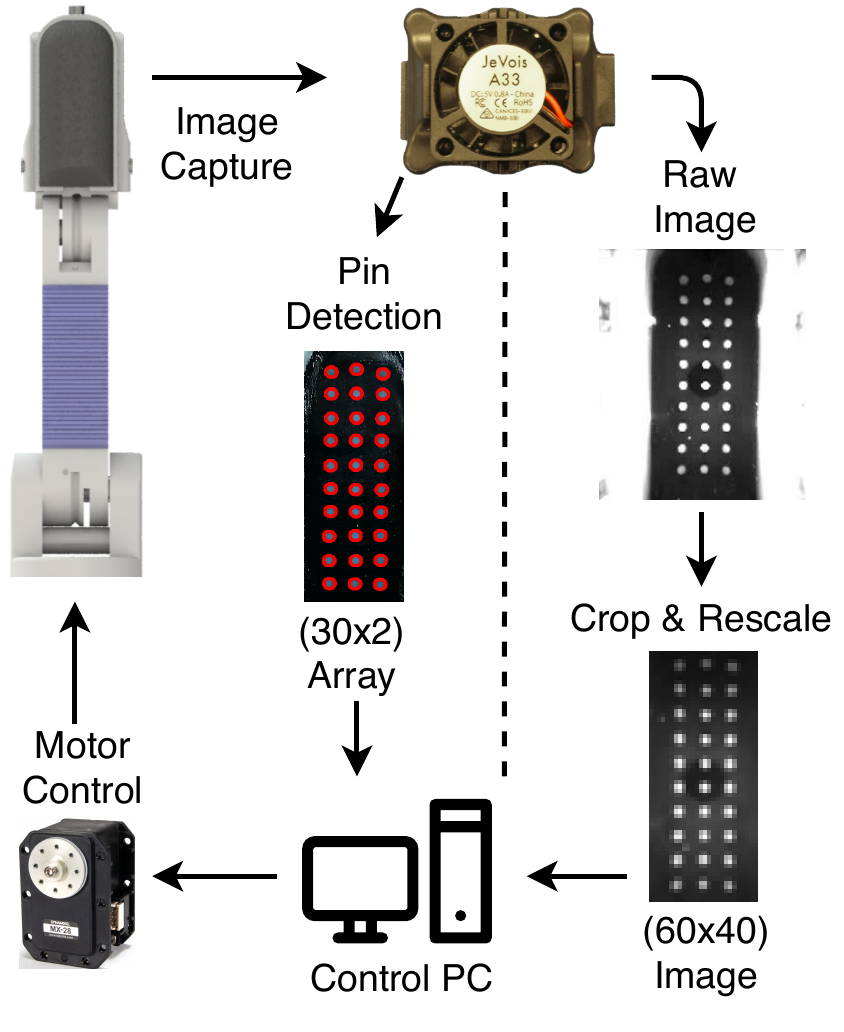}
\caption{\textcolor{black}{Flowchart showing the processes within the T-MO when images are captured by the JeVois. Either pins can be directly detected onboard the hand and sent to the control PC via a serial port or - as used here - raw images are cropped and rescaled before being passed to a classifier. Processes are the same for each of the three sensors. Note, the pixel dimension (bottom right in brackets) is an example and can be changed as required.}}
\label{fig:flow_diag}
\end{figure}

Tactile image data is collected as a video (20 FPS) while the hand performs a grasping motion. Whilst the platform allows for autonomous data collection, it still requires a significant investment of human time and effort for large datasets. Efficient use of the collected data is thus a priority.

One of the methods used to achieve efficiency is to sample multiple sequences of frames from each grasping video. To estimate the appropriate frame, we calculate the absolute pixel difference for each frame in a tactile video when compared with the first frame in the same video.\citep{Lepora2019FromSensor} This gives a basic measure of sensor deformity. 

From this method, it is possible to find the frame that corresponds with approximately $25\%$ deformation of the sensor, which we use to estimate when contact has been made with an object. We then select a sequence of 8 frames by taking every 10th frame after the initial contact frame. Repeating this with a single positive offset in the original frame number, provides additional sets from each tactile video. In total, we take 10 sets of 8 frame sequences for each grasp to increase the size of the dataset. 

Before being passed to the network some pre-processing is applied to the data, as follows (Fig. \ref{fig:flow_diag}): 
\begin{enumerate}
    \item Image cropping from the captured resolution of $320 \times 240$ down to $160 \times 220$ resolution to remove outer pixels that contain little or no tactile data
    \item Image downsampling to $40 \times 60$ resolution to retain enough of the information for the specific task while significantly reducing network size.
    \item Image concatenation horizontally over the three tactile sensors to give a $120 \times 60$ pixel image. 
\end{enumerate}

\textcolor{black}{These techniques were chosen through a combination of leveraging existing work\citep{Lepora2019FromSensor, lepora2020optimal} and author experimentation. There are lots of options that could have an effect on performance. For example, horizontal image concatenation could be replaced with vertical concatenation, stacking or using a shared weight convolutional stage for each individual sensor image;\citep{Calandra2017TheOutcomes} sharing weights between networks could improve performance by exploiting the relationship between fingers.\citep{LiuObjectRecognition}; and similarly the number of frames used per sequence could be changed or the resolution of the tactile images. We explored several of these, and the proposed method gives a simple procedure that works effectively; however, we do not rule out that other choices of setup could give slightly improved results.}


Some example images are shown in Fig. \ref{fig:frames sample}, taken from the objects used to test item classification. These images are the final frame of the extracted tactile sequences, so the sensor should be near its maximal deformation for that grasp.


\subsection{Neural Network Architecture}\label{Sec:nn}
\begin{figure}[t]
    \centering
    \includegraphics[width=0.8\linewidth]{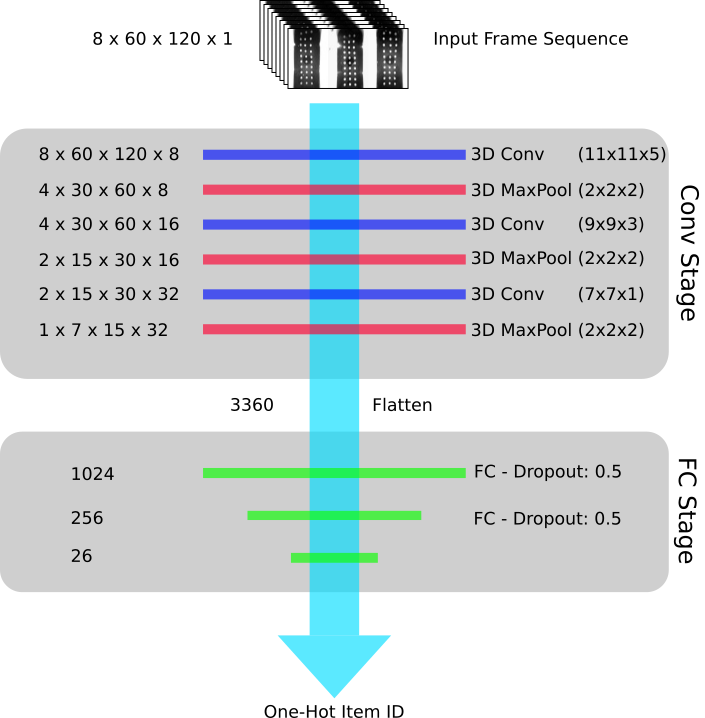}
    \caption{Convolutional network architecture used for item classification. }
    \label{fig:cnn}
\end{figure}
Recent application of Convolutional Neural Networks (CNNs) to a single TacTip has found highly robust performance for edge perception and contour following tasks,\citep{Lepora2019FromSensor} indicating the promise of deep learning for robot hands integrated with TacTip tactile sensors. As we are considering a sequence of tactile images, a $3$D CNN is appropriate to capture both spatial and temporal information.  Then the network may learn not only the geometric properties of objects being grasped but also physical properties such as compliance during the grasping process. The network architecture chosen for this task is a standard combination of convolutional layers passing forward to a fully-connected output stage (Fig. \ref{fig:cnn}).

Several methods of image augmentation are employed during training to reduce overfitting. This includes random cropping between $6\%$ and $2\%$ of width and height of an image respectively, random zooming up to an increase of $2\%$ and additive Gaussian noise with variance $\sigma_{noise}^{2} = 10^{-4}$. Random brightness and contrast adjustments are applied using $\mathrm{clip}(\alpha\times\mathrm{pixel} + \beta, 0, 255)$ where the limits for $\alpha$ and $\beta$ are randomly selected from the ranges $[0.3, 1]$ and $[-50, 50]$ respectively. These are all individually applied per frame and per sensor to best match the possible environments experienced during testing.

As a regularization technique, we use a dropout of $0.5$ on the final fully connected layers. Batch normalization, early stopping and learning rate decay are all used to improve performance. Patience values of 5 and 15 epochs are used for decaying the learning rate on a plateau and early stopping respectively. A factor of 0.25 is the value used for decaying the learning rate. The loss function used is soft-max cross entropy and the optimizer used is ADAM,\citep{Kingma2014Adam:Optimization} initialisation parameters such as learning rate are given in the subsequent sections. 

\subsection{Autonomous Grasping Platform} \label{platform}

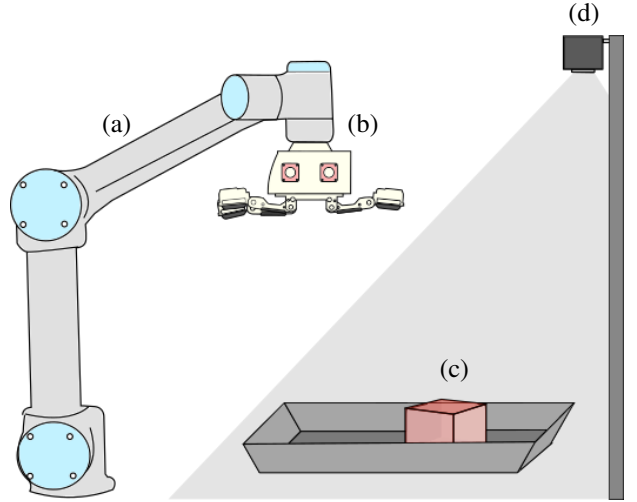
\begin{figure}[t]
\centering
    	\begin{overpic}[width=0.45\textwidth,trim={0cm 0cm 0cm 0cm},clip=true]{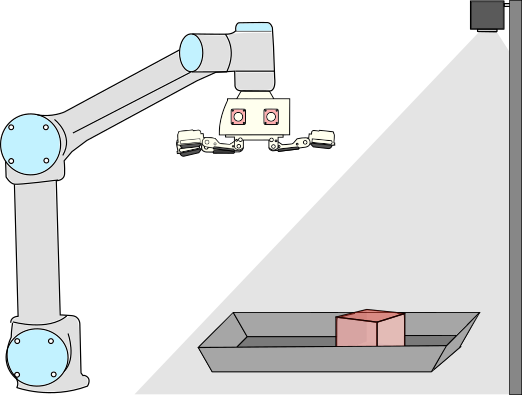}
        \put(15,60){(a)}
        \put(55,60){(b)}
        \put(70,20){(c)}
        \put(91,78){(d)}
        \end{overpic}
        \caption{Diagram of the autonomous setup used for both collecting large amounts of data and for verifying grasping capabilities of the T-MO. (a) Universal Robotics UR5 robot arm. (b) Tactile Model-O. (c)~Object to be grasped. (d) Microsoft Kinect2 RGBD camera. }
        \label{fig:setup}
\end{figure}

To facilitate testing and data collection, we have built an automated grasping platform for the hand (Fig. \ref{fig:setup}). The T-MO is mounted as the end effector on a UR5 robot arm (Universal Robots, Denmark), controlled by the PC used to capture and store the tactile data. In addition, a Kinect 2 RGBD camera (Microsoft) is mounted above the workspace to provide a depth image of objects, which we process to give putative pose estimates of use for grasping. A tray is located in front of the base of the arm into which a user can place an object and give the control PC an object label (currently the only human input into this system).

This platform uses the Kinect depth images to estimate the pose of objects, which guide a simple planner for the arm-hand system to attempt to grasp and lift test objects. \textcolor{black}{The depth image captured by the Kinect is first cropped to contain only the region covering the tray. The approximate depth of this tray is then obtained by averaging the depth across an entire image, which gives a reasonable approximation because the object occupies a relatively small area. We then consider points closer than this depth minus an offset of 10\,mm (to account for image noise) to give the 2d extent of the object.} This region is used to find the centre of mass $(x,y)$ and pose angle from the image moments, corresponding to the major and minor axes of the best-fit ellipse. A $z$ coordinate is also estimated from the minimum object depth to guide the initial vertical hand placement before grasping at 20\,mm above the object. We then transform from the coordinates in the camera frame to the robot frame to provide the position for the T-MO to grasp the object.

\begin{algorithm}[t]
	\caption{Collecting grasp data.}\label{alg: collect}
	\begin{algorithmic}[1]
		\Procedure{CollectGraspData}{\textit{objectName, numGrasps}}
		\State $\Call{Arm.Move}{\textit{homeCoords}}$
		\For{$i = 0$, $i{+}{+}$, $i < \textit{numGrasps}$}
		\State $\textit{depthImage} \gets \Call{Kinect.GetDepth}{ }$
		\State $\textit{coords} \gets \Call{GetCoords}{depthImage}$
		\State $\textit{transCoords} \gets \Call{TransformCoords}{coords}$
		\State $\Call{Arm.Move}{\textit{transCoords}}$
		\State $\textit{refImage} \gets \Call{Sensor.GetRefImage}{ }$
		\State $\textit{data} \gets \Call{Hand.Grasp}{ }$
		\State $\Call{Arm.Raise}{ }$
		\State $\textit{deformed} \gets \Call{Sensor.CheckDef}{\textit{refImage}}$ 
		\If{$\textit{deformed} = True$}
		\State $\Call{Arm.Move}{\textit{randomCoords}}$
		\State $\Call{Hand.Release}{ }$
		\State $\Call{Arm.Move}{\textit{homeCoords}}$
		\Else
		\State $\Call{Arm.Move}{\textit{homeCoords}}$
		\EndIf
		\State $\Call{SaveData}{\textit{data}}$
		\EndFor
		\EndProcedure
	\end{algorithmic}
\end{algorithm}

\begin{figure*}[t]
	\includegraphics[width=0.99\textwidth]{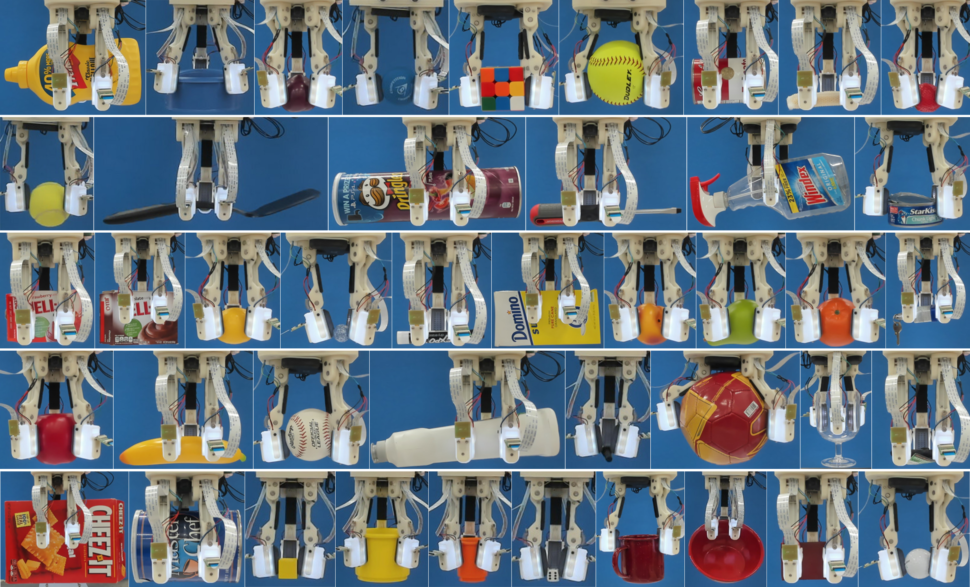}
	
	\caption{Example images of the T-MO hand grasping a variety of objects from the YCB Benchmarking Object Set. Each grasp is performed by placing the object on a flat surface in a fixed position and using a UR5 robotic arm to lower the hand over the object and raise to a specified height. This figure also shows how the rotation of the fingers can be employed to help grasp objects depending on their shape.}
	\label{fig:all_items}
\end{figure*}

Further control of the hand is available in the rotation of the finger joints. Fingers $2$ and $3$ can be jointly rotated from $0\degree$ (Cylindrical grasp), through $45\degree$ (Spherical grasp), to $90\degree$ (Opposed grasp). In Section \ref{grasping validation} the grasp that is likely to be successful is then manually selected from these distinct cases depending on the geometry of the object to be grasped ({\em e.g.} Fig.~\ref{fig:all_items}). In Sections \ref{item_classification} and \ref{grasp_success} we define an automatic procedure to choose the rotation of the finger joints. To do this we use the ratio of major to minor axis detected by the outline of the object using the depth image from the Kinect. If an object has a $1:1$ axis ratio it is assumed that a spherical grasp ($45\degree$) will be best. Similarly if an object has an aspect ratio of $1:3$ or higher it is assumed a cylindrical grasp ($0\degree$) will be best. Overall, we found that this mapping of aspect ratio to $[0\degree,45\degree]$ joint angle worked best for the subset of objects used here.

The tactile sensors have an immediate use within this autonomous platform, as they indicate both grasp success and also which fingers contact a held object. After the hand has grasped and attempted to lift an object, the tactile images from the three fingers at the peak of the raising movement are compared to initial non-deformed reference images prior to grasping. \textcolor{black}{Previous work has demonstrated that the Structural Similarity Index Metric (SSIM) \citep{Wang2004ImageSimilarity} can provide an accurate difference metric for tactile images.\citep{lee2019touching} We use SSIM to measure the difference between each tactile image and its non-deformed reference:}

\begin{equation}
\label{eq:ssim}
\mathrm{SSIM}(u,v) = \frac{(2\mu_{u}\mu_{v} + c_1)(2\sigma_{uv} +c_2)}{(\mu_{u}^2 + \mu_{v}^2 +c_1)(\sigma_{u}^2+\sigma_{v}^2+c_2)}
\end{equation}
\noindent where $u$ and $v$ represent a window of $N\times N$ pixels (here $N=7$) within the two images to be compared, $\mu$ and $\sigma$ represent the mean and covariance of the windows given, and $c_1$ and $c_2$ are regularizing constants defined to stabilize the division. The regularizing constants are calculated with $c_1 = (k_1L)^2$ and $c_2 = (k_2L)^2$ where $k_1=0.01$, $k_2=0.03$ and $L$ is the dynamic range of the pixel values (here $255$). A single numeric similarity value for each sensor is calculated by averaging this metric over a sliding window taken across the entire image. If these values are below a predefined threshold (here $0.96$) for two or more sensors, then the grasp is considered a success. 

For more robust detection, this process is repeated over 20 frames and the mode SSIM  taken as a final measure. The SSIM metric is chosen because we expect it to be robust to lighting changes likely to occur during movement of the hand (which can effect the tactile image). Testing showed that this was a reliable method and produced minimal false positive or negative results. The full method for data collection is given in Algorithm \ref{alg: collect}.

\begin{figure*}[t]
	\centering
	\begin{subfigure}{0.375\textwidth}
		\includegraphics[height=4.2cm,left]{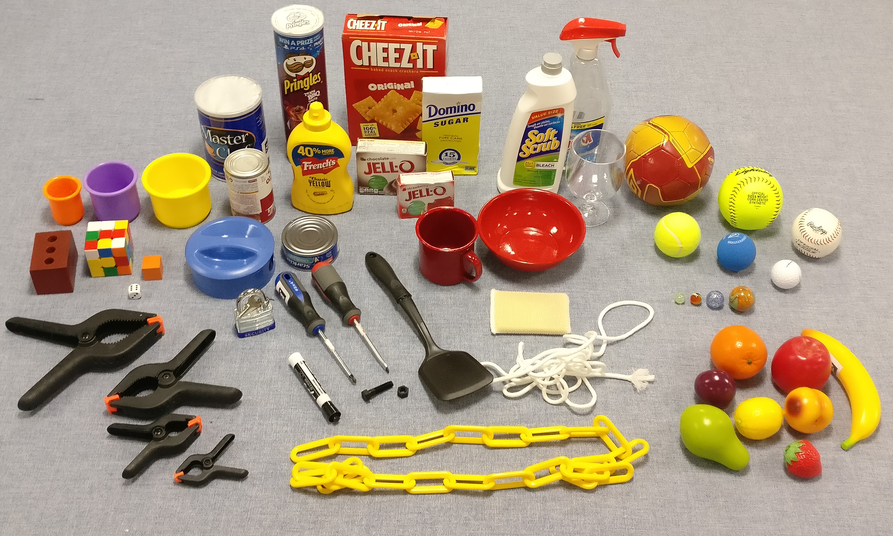}
		\caption{}
	\end{subfigure}
	\centering
	\begin{subfigure}{0.3\textwidth}
		\includegraphics[height=4.2cm,right]{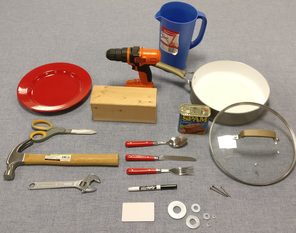}
		\caption{}
	\end{subfigure}
	\begin{subfigure}{0.31\textwidth}
		\includegraphics[height=4.2cm,right]{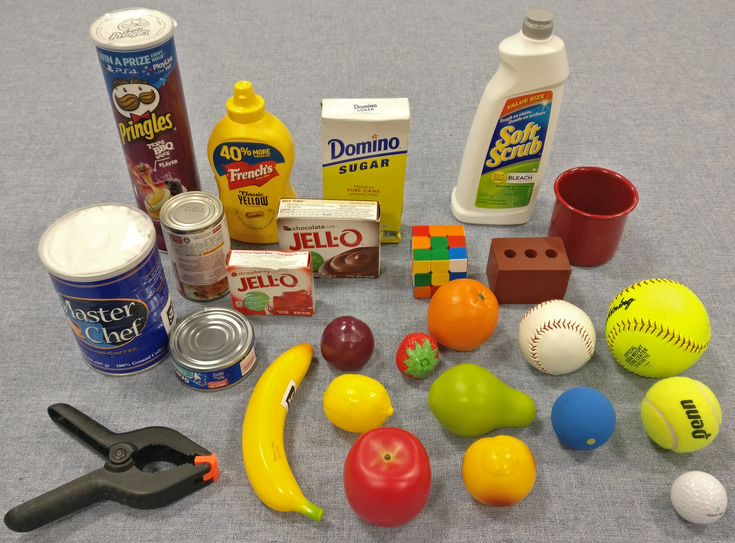}
		\caption{}
	\end{subfigure}
	\caption{(a) Items from the YCB object set that the T-MO was able to grasp. (b) Items from the YCB object set that the T-MO was unable to grasp. (c) Objects from the YCB object set used in the classification task.}
	\label{fig:items_grasp}
\end{figure*}

\section{Experiment Description}
\subsection{Grasping Benchmarks}\label{grasping validation}
\textcolor{black}{First, we assess how modifying the existing grasping system of the Model O hand to include tactile sensors affects grasping performance, considering four grasping benchmarks from other studies. These are mainly based on Yale-CMU-Berkeley (YCB) object set, which comprises 77 objects in 5 categories including food items and tools.\citep{Calli2015BenchmarkingSet}}

\textit{Grasping Benchmark 1 - Complete YCB Set:} To start, we test every object in the YCB object set to determine whether it can be grasped by the T-MO. \textcolor{black}{Each object was placed at a fixed position directly beneath the hand, which was then lowered to a predefined height, a grasp performed, and the object lifted. Both the set height that the hand was lowered to and the grasp type (Opposed, Cylindrical or Spherical) were pre-specified on a per object basis.}

\textit{Grasping Benchmark 2 - i-HY Comparison:} The Model O is hand is originally based on the i-HY hand, which demonstrated its grasping capabilities with a benchmark involving 10 unique objects grasped 20 times.\citep{odhner2014compliant} Each grasp was automated by detecting the approximate centre of an object using the depth map from a Kinect RBGD sensor and aligning to the major axis of the detected object. The same experiment is performed here (details in Section \ref{platform}), but as the objects in the i-HY study were not standardized, we have used the most similar objects we could find.

\textit{Grasping Benchmark 3 - Gripper Assessment Benchmark:} Another official benchmark is the Gripper Assessment Benchmark, provided by \citet{Calli2015BenchmarkingSet}, which is performed with a subset of objects from the YCB set. This benchmark does not use of the depth sensor but instead a fixed position as described in the benchmarking guidelines.

\textit{Grasping Benchmark 4 - Extended Gripper Assessment Benchmark:} An extension to the standard Gripper Assessment Benchmark was proposed by \citet{Jamone2016BenchmarkingSet} to include three more categories of objects: Cubic, Cylindrical and Complex. The addition of further objects provides more information about types of object graspable by the T-MO. The experiment is undertaken in the same manner as Benchmark 3 above.

\subsection{Tactile Sensing Test \rom{1}: Item Classification}\label{item_classification}

Bearing in mind that our modification of the Model O design was to use tactile sensing for improving the hand's functionality, we first validate the tactile sensing quality by classifying objects using the tactile images alone.

For this test, we chose a subset of $26$ items from the YCB object set that satisfy two properties (Fig. \ref{fig:items_grasp}c): they must be consistently graspable with the hand and be clearly detectable with the depth sensor (which struggles with small or clear objects). For added difficulty, some of the objects have been chosen to share either global or local geometric similarities. 

To grasp the objects, we rotate the fingers of the T-MO to $[0\degree,45\degree]$ using only the spherical and cylindrical grasps (Section \ref{platform}). The opposed grasp at $90\degree$ rotation is omitted because it works best with very small objects, which are not present in the objects chosen for this tactile task. For each of the $26$ objects, data was collected for $20$ grasps, giving $520$ unique grasp videos. \textcolor{black}{The object position was varied for each grasp by randomly moving the object after a grasp and dropping from a small height of 5cm. This exposed different object orientations and therefore different tactile responses during data collection. It was necessary to ensure some objects faced up otherwise they were not graspable with our simple top-down grasp planner; for example, long cylindrical objects were placed with their elongated side on the tray surface.} 

The processed tactile data is then fed into a network trained to classify these objects. The architecture is as described in the Methods, but with learning rate initialised at $10^{-4}$ and weights randomly initialised from a Gaussian with mean $0$ and standard deviation $0.01$. The collected data is separated into training and validation sets with a $70\%$ to $30\%$ split on a per video basis (not a per sequence basis) to avoid training and testing on sequences of frames from the same grasp video.

An online experiment is used to further validate the performance on distinct test data, with $4$ grasps performed on each of the 26 objects. Multiple tactile image sequences are taken from the grasp video and fed into the trained network as a batch (Section~\ref{tactiledata}). The final prediction is the $\argmax$ of the mean prediction over the batch; in effect, the most confident prediction over all sequences extracted from a single grasp.

\subsection{Tactile Sensing Test \rom{2}: Grasp Success Prediction} \label{grasp_success}

Next, we use touch to predict whether an established grasp will be successful when lifting an object. The data is labelled (success/failure) using the SSIM-based measure (Section~\ref{platform}) of whether the fingers are still in contact after a lift attempt.

To collect a balanced dataset with both failed and successful grasps on the same object, the tactile data was collected while introducing random perturbations in the hand pose and grasp. The hand pose was perturbed by a uniform-random distributed $[-20,20]$\,mm variation in the $(x,y)$-coordinates, a $[0,20]$\,mm $z$ perturbation and a $[-30\degree, 30\degree]$ axial perturbation. The finger joint rotation was also varied randomly between $[0\degree,45\degree]$ and the maximum torque between $20\%$ to $35\%$ of the maximum instantaneous and static motor torque (compared with $30\%$ as previously used), which also affected the speed of finger movement during a grasp. Overall, these perturbations reduced the grasp success to $80\%$ of the $520$ collected grasps (Fig.~\ref{fig:all_items}; same objects as Section~\ref{item_classification}).

The processed tactile data was again fed into a neural network, with the same architecture as in Section~\ref{item_classification}. Minor hyperparameter changes (dropout increased to $0.75$; learning rate decreased to $10^{-6}$) are made to reduce overfitting, which became more prevalent in this task. The output layer gives a two-label classification with each grasp predicted as a success or failure according to the mean output of the final softmax layer of the network (successful grasp category being $>0.5$). 

\textcolor{black}{A final aspect of this experiment measures whether there is a correlation between force applied when grasping an object and grasp success prediction. We performed an additional experiment in which the maximum motor torque is varied over 12 grasps from barely touching to firmly holding the object (from $11\%$ to $35\%$ of the maximum instantaneous and static motor torque, where each grasp has an increment of $2\%$ in this torque). This test was performed over four objects having uniform grasping surfaces (Baseball, Orange, Bleach, GumPot) to control for factors such as positions of edges and textures on more complex objects. The GumPot is a novel object to test if the learned features generalize outside of the objects used for training.
}
\begin{figure}[t]
\centering
\includegraphics[width=0.9\linewidth]{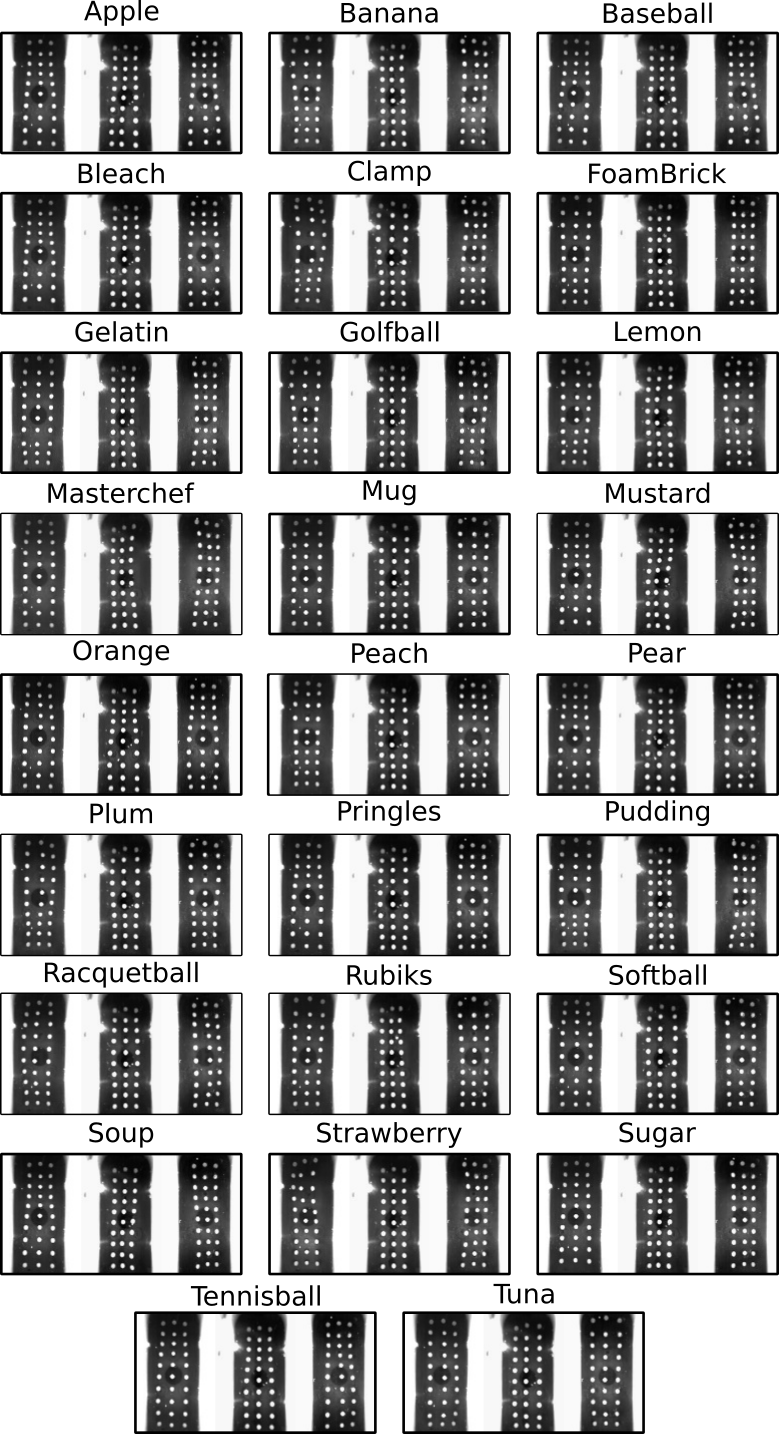}
\caption{Examples of tactile images extracted from a grasp for each object used in the item classification task. Each image is taken when the sensor is under its maximal deformation. \textcolor{black}{The bright spot visible at the top left of each middle sensor image is caused by glare from the LEDs on the sensor's acrylic lens. As this is consistent across all objects the glare does not provide any useful features from which a CNN can learn; therefore, we do not expect this to have an effect on tactile sensing.}}
\label{fig:frames sample}
\end{figure}
\subsection{Tactile Sensing Test \rom{3}: Sensitivity Analysis} \label{sec:sensitivity}

Our final analysis investigates how the T-MO behaves with a reduced number of tactile sensors. \textcolor{black}{All systems are liable to breakage or temporary inaction, so determining whether a system remains effective when operating at reduced capacity is an important consideration.} This test also gives an improved understanding of how the tactile information has been employed for item classification and grasp success prediction.

\textcolor{black}{In this test, we re-used the data from the previous experiments of item classification (Test I, Section~\ref{item_classification}) and grasp success prediction (Test II, Section~\ref{grasp_success}) with all possible combinations of `working' tactile sensors. The validation accuracy is then assessed after retraining the neural networks using these limited tactile observations. }


\begin{table}[b]
	\centering
	\caption{\label{YCB Grasping Benchmark}Scoring of the Tactile Model O hand at four set points (SP) in the Basic Gripper Assessment Benchmark provided by \citet{Calli2015BenchmarkingSet}. F.O. and A.O. stand for Flat Objects and Articulate Objects. Total score for this test is 202.5 out of 404. }
	
	\begin{tabular}{|cl|c|c|c|c|}
		\hline
		& \multicolumn{1}{c|}{ \textbf{Object} } & \multicolumn{1}{c|}{ \textbf{SP1} } & \multicolumn{1}{c|}{ \textbf{SP2} } & \multicolumn{1}{c|}{ \textbf{SP3} } & \multicolumn{1}{c|}{ \textbf{SP4} } \\ \hline \hline
		
		\multirow{9}{*}{\STAB{\rotatebox[origin=c]{90}{ \textbf{Round Objects} }}}
		& Soccer Ball   & 4 & 4 & 4 & 4   \\
		& Soft Ball     & 4 & 4 & 4 & 4   \\
		& Tennis Ball   & 4 & 4 & 4 & 4   \\
		& Racquet Ball  & 4 & 4 & 4 & 4   \\
		& Golf Ball     & 4 & 4 & 4 & 4   \\
		& Marble XL     & 4 & 4 & 4 & 4   \\
		& Marble L      & 4 & 4 & 0 & 4   \\
		& Marble M      & 4 & 0 & 4 & 0   \\
		& Marble S      & 4 & 4 & 0 & 0   \\ \hline \hline
		
		\multirow{2}{*}{\STAB{\rotatebox[origin=c]{90}{ \textbf{F.O.} }}}
		& Washer 1-6   & 0 & 0 & 0 & \cellcolor{lightgray} \\
		& Credit Card  & 0 & 0 & 0 & \cellcolor{lightgray} \\  \hline \hline
		
		\multirow{9}{*}{\STAB{\rotatebox[origin=c]{90}{ \textbf{Tools} }}}
		& Marker L          & 4 & 0 & 4 & 0   \\
		& Scissors          & 0 & 0 & 0 & 0   \\
		& Hammer            & 0 & 0 & 0 & 0   \\
		& Flat Screwdriver  & 2 & 0 & 0 & 0   \\
		& Drill             & 0 & 0 & 0 & 0   \\
		& Clamp XL          & 4 & 4 & 4 & 4   \\
		& Clamp L           & 4 & 4 & 4 & 4   \\
		& Clamp M           & 4 & 4 & 4 & 4   \\
		& Clamp S           & 4 & 4 & 4 & 0   \\ \hline \hline

		\multirow{2}{*}{\STAB{\rotatebox[origin=c]{90}{ \textbf{A.O.} }}}
		& Rope          & 8.5 & \multicolumn{3}{c}{\cellcolor{lightgray}} \vline \\
		& Chain         & 0   & \multicolumn{3}{c}{\cellcolor{lightgray}} \vline \\ \hline
	\end{tabular}
\end{table}

\begin{table}[b]
	\centering
	\caption{\label{YCB Grasping Benchmark Extension}Extended scoring of the Tactile Model O hand in the Basic Gripper Assessment Benchmark as proposed by \citet{Jamone2016BenchmarkingSet}. Total score for this test is 176 out of 208. }
	\begin{tabular}{|cl|c|c|c|c|}
		\hline
		& \multicolumn{1}{c|}{ \textbf{Object} } & \multicolumn{1}{c|}{ \textbf{SP1} } & \multicolumn{1}{c|}{ \textbf{SP2} } & \multicolumn{1}{c|}{ \textbf{SP3} } & \multicolumn{1}{c|}{ \textbf{SP4} } \\ \hline \hline
		
		\multirow{4}{*}{\STAB{\rotatebox[origin=c]{90}{ \textbf{Cubic} }}}
		& Pudding Box          & 4 & 4 & 4 & 4   \\ 
		& Foam Brick           & 4 & 4 & 4 & 4   \\ 
		& Coloured Wood Block  & 4 & 4 & 0 & 0   \\ 
		& Dice                 & 4 & 4 & 0 & 0   \\ \hline \hline
		
		\multirow{5}{*}{\STAB{\rotatebox[origin=c]{90}{ \textbf{Cylindrical} }}}
		& Pringles Can       & 4 & 4 & 4 & 4   \\ 
		& Plastic Wine Glass & 4 & 4 & 4 & 4   \\ 
		& Cup L              & 4 & 4 & 4 & 4   \\ 
		& Cup M              & 4 & 4 & 4 & 4   \\ 
		& Cup S              & 4 & 4 & 4 & 4   \\ \hline \hline
		
		\multirow{4}{*}{\STAB{\rotatebox[origin=c]{90}{ \textbf{Complex} }}}
		& Plastic Pear       & 4 & 4 & 4 & 4   \\ 
		& Plastic Strawberry & 4 & 4 & 4 & 4   \\ 
		& Plastic Bolt       & 4 & 4 & 0 & 0   \\ 
		& Plastic Nut        & 4 & 4 & 0 & 0   \\ \hline 
		
	\end{tabular}
\end{table}

\section{Results}
\subsection{Grasping Benchmarks}
The T-MO successfully grasped the majority of the objects in the YCB set (Grasping Benchmark 1) only struggling on the larger heavier objects and the smaller flat ones (successful/unsuccessful objects shown in Fig.~\ref{fig:items_grasp}). 

The T-MO also had strong results on 7/10 objects based on those used to test the i-HY (Grasping Benchmark 2), with at least $18$ out of $20$ trialled grasps being successful. The remaining 3 objects - a hammer, file and pen - were unsuccessful because the thickness of the redesigned distal phalanges did not allow us to use the previous technique of driving the hand into the table to perform a power grasp. Similarly the compliance of the tactile sensors results in pinch grasps being more subject to failure when given high torsional forces. \textcolor{black}{Although the i-HY apparently scored better than the T-MO in this test (the i-HY only struggled with the pen), there are thus subtleties with comparing these results directly. As the objects tested on the i-HY were not standardised, we could only use similar but not identical objects in our comparison. Nevertheless, this test does indicate the physical properties of objects that the T-MO is likely to struggle with, indicating limitations in the design.} 

To further validate the grasping capabilities of the T-MO hand, two standardised benchmarks, the Gripper Assessment Benchmark (GAB) and Extended Gripper Assessment Benchmark (EGAB) were deployed. \textcolor{black}{These two tests carry much more weight when comparing to other systems as they involve a rigorous, standardised procedure.} Overall, the total score for the GAB was $202.5$ out of $404$ (complete scores for each object shown in Table~\ref{YCB Grasping Benchmark}), and the total score for EGAB was $176$ out of $208$ (complete scores in Table \ref{YCB Grasping Benchmark Extension}).

Therefore, performance of the T-MO hand is competitive with other hand designs, surpassing the iCub (173/404) and Model T (122/404) scores. The T-MO also surpassed the iCub score (164/208) in the Extended Benchmark. That said, flat objects prove a significant challenge for this hand, resulting in the drop in the overall score. Whilst it would be possible to modify this hand with a nail design to help grasp flat objects, this would require significant modification of the tactile sensing part of the design to remain effective. To guide this modification, it would be useful to also have a comparison with the original Model-O hand, but to the best of our knowledge this is not available.

Other GRAB Lab robotic hands do achieve better scores on the standard benchmark. These include the Model B, Model T42 and Model S, which score 270, 379 and 402.5 out of 404 respectively (none of which have tactile sensing). These higher scores are mostly due to the fingernail designs useful for picking up flat objects and the ability to pick up heavy objects using a power grasp. Most tactile sensors focus on the fingertips of robotic hands, avoiding fingernail designs and palm sensors. This generally reduces the ability to grasp flat and heavy objects and results in lower benchmark scores. To the best of our knowledge, the T-MO is the highest scoring hand with tactile capabilities. 

\subsection{Tactile Sensing Test \rom{1}: Item Classification}\label{test1}
\begin{figure*}[t]
	\centering
	\includegraphics[width=1.0\linewidth]{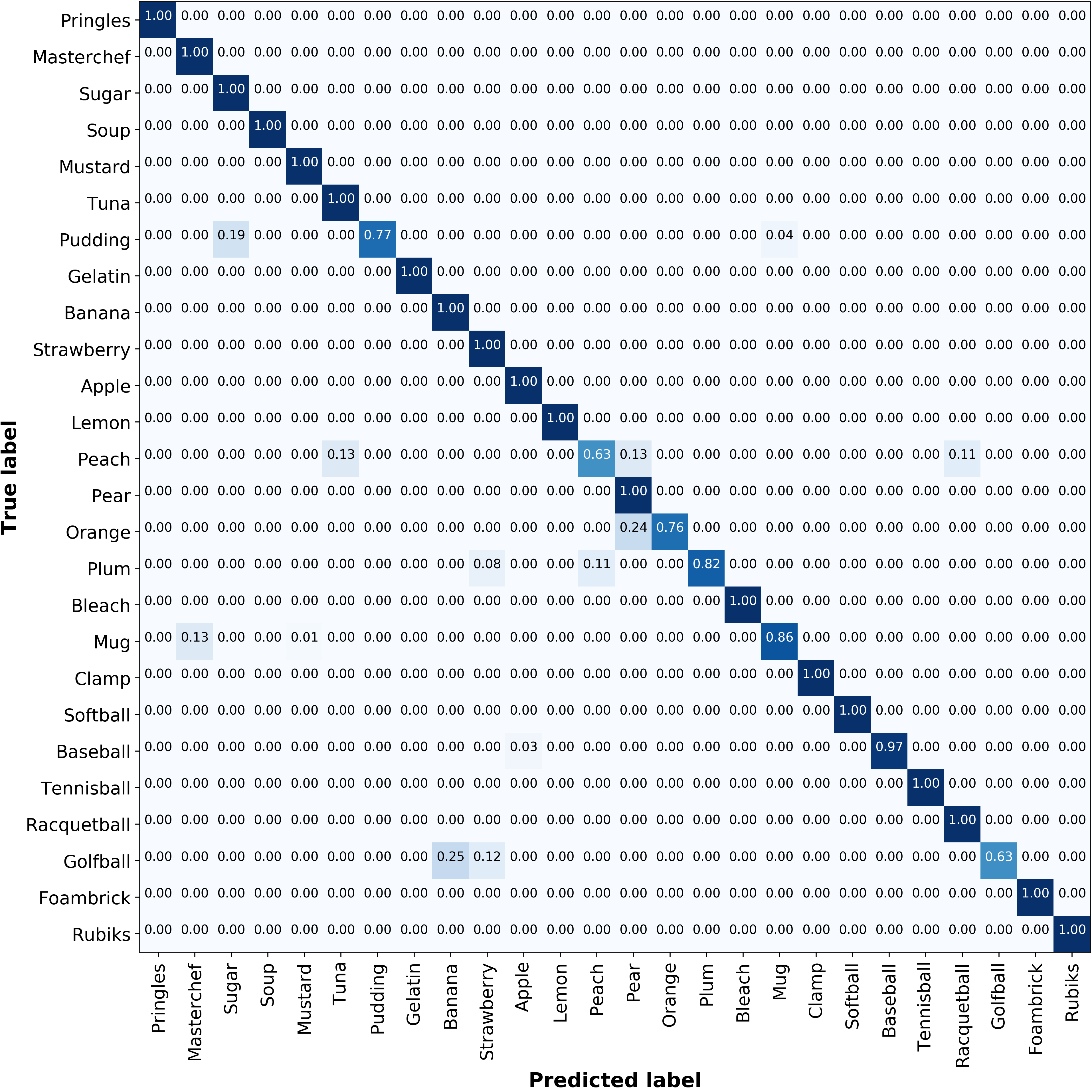}
	\caption{Confusion matrix demonstrating the validation results of the object classification based on purely tactile information task. The overall validation accuracy for this task is approximately 93\%.}
	\label{fig:cnf mtrx}
\end{figure*}

The first test of the tactile sensing capabilities involved identification of $26$ objects (Fig.~\ref{fig:items_grasp}(c)) from a grasp using only tactile data fed into a CNN (example tactile image for each object shown in Fig.~\ref{fig:frames sample}). The overall accuracy found over this set was $93\%$. Misclassifications tend to be on objects that share some geometric similarities (confusion matrix shown in Fig.~\ref{fig:cnf mtrx}). For example, the pudding box and sugar box are both rectangular with similar depths (32\,mm and 37\,mm), giving an edge in the sensor images at approximately the same location. 

For round objects objects, the misclassifications seem reasonble; for example, depending on the grasp direction, the peach and pear objects are of similar size and curvature. \textcolor{black}{The worst performance for item classification was on the golf ball. We attribute this to it being the smallest object which may lead to frames being passed to the CNN prior to any sensors contacting the object. Consequently, there are fewer tactile images in which deformation occurs resulting in less data from which the networks can learn useful features.}

When testing on online test data, the overall accuracy was $77\%$ with similar misclassifications as the offline validation described above. Considering there are 26 objects many of which have geometric similarities, this is good performance but clearly poorer than the $93\%$ validation accuracy. \textcolor{black}{This drop in performance indicates that the training has not generalized fully to off-sample data, most likely because there is a large number of possible positions and orientations for unconstrained objects}. We expect a larger training set would enable better determination of features that generalise across different positions and orientations. The present data set is sufficient, however, to show reasonable online performance and indicate directions for future improvement. 

\subsection{Tactile Sensing Test \rom{2}: Grasp Success Prediction}\label{test2}

The second test of the tactile sensing was to predict whether a grasped object, once lifted, would be successfully held. On successful grasps, the classifier was almost always correct with 98\% true positives, and only 2\% of false positive predictions when the grasp was predicted to fail but was actually a success. On unsuccessful grasps, 82\% true negatives were correctly predicted, with 18\% false negatives of predicting a grasp would succeed when it actually failed (Table \ref{tab:grasp_cnf_mtrx}). 

\begin{table}[h]
	\centering
	\caption{\label{tab:grasp_cnf_mtrx}Confusion matrix demonstrating the validation accuracy of the trained neural network predicting whether a grasp will be successful or unsuccessful.}
	\bgroup
	\def\arraystretch{1.25}
	\begin{tabular}{c|cc|c}
		& \multicolumn{2}{c|}{\textbf{Predicted}} &  \\ \hline
		Failure & $\frac{304}{373}\approx$ \textbf{82\%} & $\frac{69}{373}\approx$ 18\%  & \multirow{2}{*}{\STAB{\rotatebox[origin=c]{270}{\textbf{True}}}} \\
		Success & $\frac{23}{1483}\approx$ 2\%  & $\frac{1460}{1483}\approx$ \textbf{98\%} &  \\ \hline
		& Failure & Success &   \\ 
	\end{tabular}
	\egroup
\end{table}

However, the baseline accuracy for this task is that $80\%$ of all grasps were successful (20\% unsuccessful). Therefore, overall the predictor correctly predicted true positives or true negatives on $95\%$ of the validation data (since the 18\% false negative rate was on 20\% of the data). It is evident that the bias prevalent in the data, which contains fewer unsuccessful grasps, is also present in the trained network, with the majority of misclassifications being false negatives.

Based on our observations of failure cases, we hypothesise that the motor torque variation was having the most effect on grasp success prediction, which makes sense intuitively because less firmly held objects will provide poorer tactile data. To test this hypothesis, we performed an additional experiment in which the maximum motor torque is varied over grasps from barely touching to firmly holding the object.

\begin{figure*}[t]
	\centering
	\begin{subfigure}{0.45\textwidth}
        \includegraphics[width=0.9\linewidth]{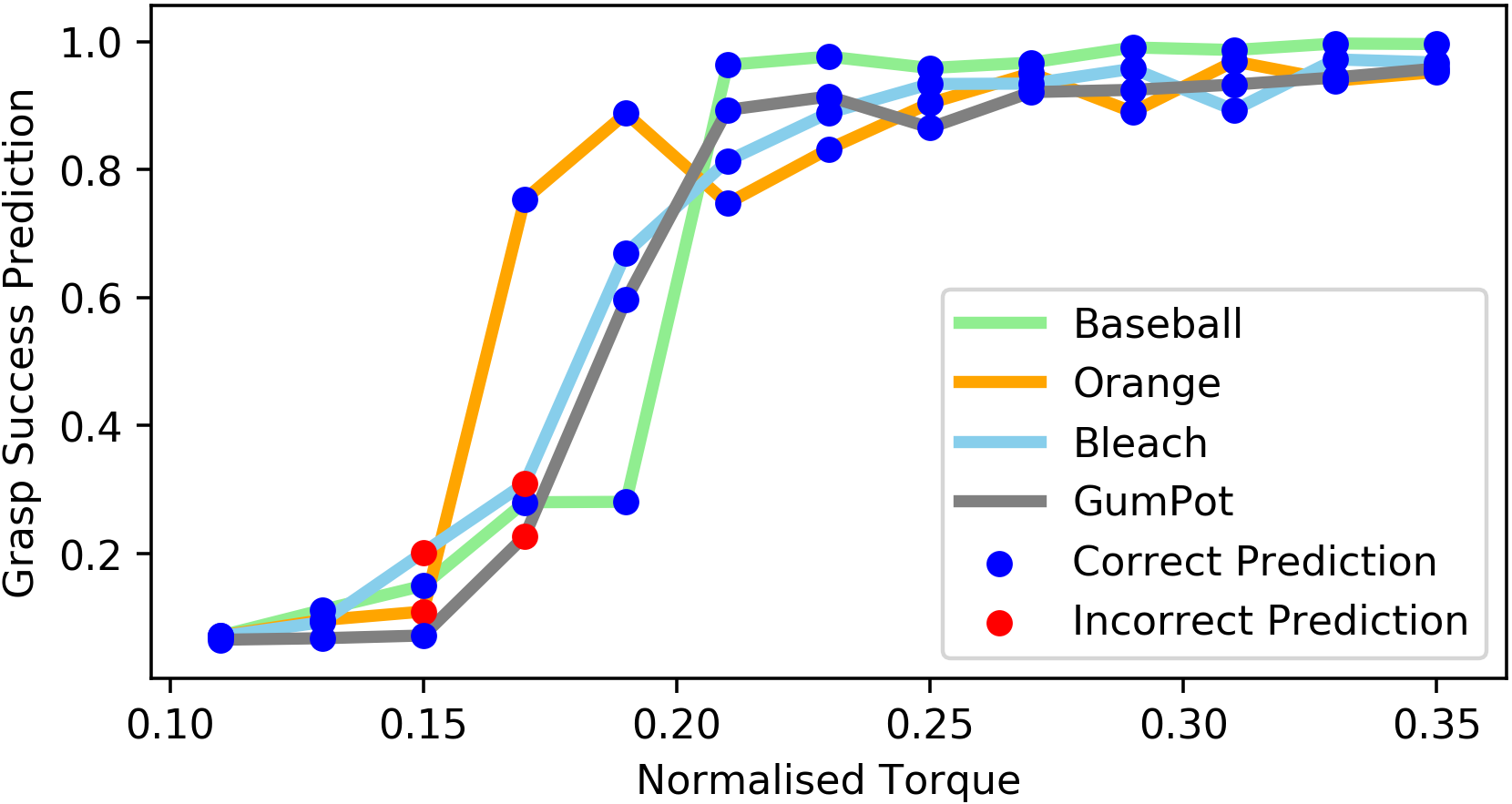}
		\caption{}
	\end{subfigure}
	\centering
	\begin{subfigure}{0.49\textwidth}
    	\includegraphics[width=1.0\linewidth]{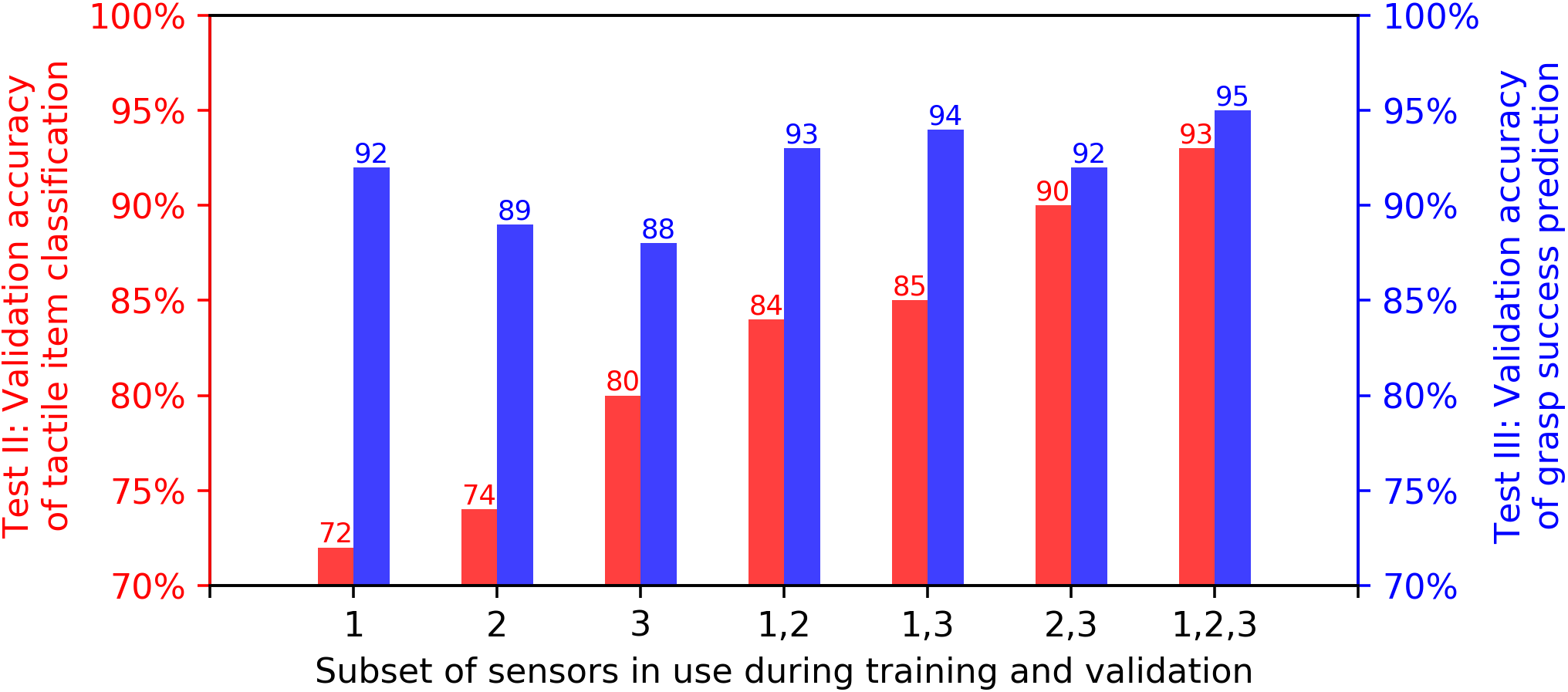}
		\caption{}
	\end{subfigure}
	\caption{(a) Variation in output of the grasp prediction network with applied torque.(b) Performance in both the item classification and success prediction tasks whilst using tactile information from a subset of the available sensors; in both cases, best performance is achieved when using all three sensors, although strong performance can also be achieved with only two active sensors.}
	\label{fig:torque_variation}
\end{figure*}

The results of this test indicate a strong trend between applied force and successful grasp prediction (Fig.~\ref{fig:torque_variation}(a)). Once the grasp has been established and the prediction made, the arm attempts to raise the object and a ground truth of whether the grasp was successful or not is taken (Fig.~\ref{fig:torque_variation}(a); blue/red points for correct/incorrect predictions). The overall accuracy achieved in this test matched that of the validation data with $92\%$ of grasps being classified correctly ($75\%$ for predicting unsuccessful grasps; $100\%$ for successful grasps). 

Overall, the network has learnt that a higher grasping force is more likely to be successful. There is a sharp drop where the network switches from predicting successful to unsuccessful grasps (near a normalized torque of 0.19). Torques close to this boundary are where incorrect predictions are most common (red markers, Fig.~\ref{fig:torque_variation}(a)). High and low grasping force tends to result in correct predictions, which we interpret as from either the tactile information being sufficiently good for accurate prediction (high torque) or because the classifier indicated an insufficient torque for a successful grasp.

\subsection{Tactile Sensing Test \rom{3}: Sensitivity Analysis}
	
To determine how well the T-MO performs in scenarios where some sensors are not operational, our final test repeated the previous two experiments with various sensor combinations. As expected, the best performance for both tests was when using all three tactile sensors (Fig.~\ref{fig:torque_variation}(b)). Also as expected, the performance dropped for all 3 combinations of 2 sensors (sensors 1,2; 1,3; and 2,3) and dropped once again when using only of 1 of the 3 tactile sensors (sensors 1, 2 or 3). However, the drop in performance reduction from 3 to 2 sensors was relatively small, with a mean drop of $7\%$ from 93\% for Test I and $2\%$ from 95\% for Test II, revealing redundancy in the sensor readings. The performance drop was larger from three to a single sensor (mean 18\% for Test I and 5\% for Test II), but perhaps not as much as anticipated. 

In Test II (Section~\ref{test2}), the highest single sensor accuracy for grasp success prediction is for sensor $1$. This is expected because this finger is needed to apply a force that opposes the other two fingers during all successful grasps, and therefore should carry the most relevant information to indicate a failing grasp. However, in Test I (Section~\ref{test1}), the opposite trend is visible: sensor 1 scores the lowest accuracy for a single sensor. To explain this, we observe that sensor 1 is aligned to the major axis of the object during grasping (Section~\ref{platform}), whereas the other two fingers are rotated depending on the object. Therefore, there may be less variation for sensor 1 on the object surface than the other two sensors, and hence less information about object identity. 

Whilst there is a significant reduction in performance when using only a single sensor, particularly on the item classification task, an interesting area for further study is how to combine the predictions from all three single-sensor networks. Single-sensor networks have significant hardware benefits, because they enable individual predictions using the embedded GPU functionality for each tactile sensor.

Therefore, for a final sensitivity test, we make a prediction voted by networks trained for each individual sensor (rather than having all three images fed into the same network). A prediction is made by taking the mean of all values predicted by the three networks. When applying this technique to the validation data, the performance improves to $91\%$ in the item classification task (from mean 75\%), close to the 93\% performance of a network trained directly on all three sensors. Similarly, the performance on the grasp success prediction task is $92\%$ (from mean 90\%) compared with 95\% when combining all three tactile images, indicating the benefit of using individual tactile sensor predictions.


\section{Discussion}
We have presented a three-fingered tactile hand that comprises a GRAB Lab Model O modified to include three TacTip soft biomimetic optical tactile sensors in its fingertips. Using small camera modules mounted in the the distal phalanx of each finger coupled with the JeVois vision system housed in the `palm' of the hand and lightweight TacTip sensors, provided an integrated and sophisticated tactile sense that complements the grasping functionality of the hand.

To evaluate the capabilities of the tactile Model O (T-MO), we benchmarked its grasping performance using the Gripper Assessment Benchmark on the YCB object set.\citep{Calli2015BenchmarkingSet} We then tested the tactile sensing capabilities with two experiments: firstly, tactile object classification and secondly, predicting whether a grasp will successfully lift an object. 

\subsection{Grasping Capability}
In the Gripper Assessment Benchmark we scored similarly (202.5/404) to the Model T and iCub hands (122/404 and 173/404) but were surpassed by other hands produced by the GRAB Lab.\citep{Dollar2019YCBResults} This included a test where the iHY hand (on which the model O is based) could grasp all 20 objects and the T-MO could reliably grasp 17 items. The reason for this performance drop is the change to the finger morphology needed to integrate our tactile sensors within the distal phalanx that can be improved in future design iterations of the hand. Due to the thickness of the redesigned distal phalanges, the technique of driving the hand into the table in order to perform a power grasp was no longer possible.

Future design improvements to the tactile fingertips will focus around a more compact design that would help grasp small objects with a `fingernail' and larger objects by sliding under them to establish a power grasp. At present, to avoid distortion of the tactile image we use a lens with a 90\degree  field-of-view, which constrains the camera to be mounted fairly high above the sensing surface. One redesign strategy would be to adopt mirrors that enable the camera to be mounted nearer the joint, as in the GelSlim.\citep{Donlon2018GelSlim:Finger} That said, camera technology is going through a stage of rapid miniaturization, and it may be a combination of smaller camera with a wide-angle lens would be more effective. \textcolor{black}{Additionally, using different gel mediums and skin materials to test the effect of sensor friction and compliance on grasping capability would be an interesting future study. For example, a higher friction surface would allow for objects to be grasped with lower force and a more compliant sensor may improve tactile perception because of its greater deformation}, which are subjects for future study.

\subsection{Tactile Sensing}
When attempting an item classification task (Test I: Section~\ref{test1}), we obtain $93\%$ validation accuracy on 26 objects randomly placed in the tray to vary the grasp. This performance is comparable to other studies, such as \citet{Spiers2016Single-GraspSensors} ($94\%$), \citet{Schmitz2014TactileDropout} ($88\%$) and \citet{FunabashiObjectRecognition} ($95\%$) but is surpassed by \citet{Flintoff2018Single-GraspSensors} ($99\%$). That said, a direct comparison between studies is not possible because the hands and tactile sensors differ, along with the objects and experiments. In particular, our T-MO picked objects off a table, whereas other studies such as \citet{regoli2017controlled} obtain up to $98\%$ on a smaller set of 21 objects that were passed to static mounted hands, which does not control against the human help provide a better grasp for object classification.

Our other main test was to predict whether a grasp would successfully lift an object (Test II: Section~\ref{test2}). We obtain 95\% accuracy of grasp success prediction on the same 26 objects as Test I but with their grasp poses perturbed randomly so that some grasps fail upon lifting. \citet{Calandra2017TheOutcomes} performed a similar experiment and obtain $75.6\%$ validation accuracy when using tactile information from two GelSight sensors. This was performed on a larger dataset with more (and different) objects, and the data was split such that an object was only in the training or test set, so again a direct comparison is not possible. \textcolor{black}{We also exceed the results of \citet{WanVariabilityGrasping} (90\%) and \citet{krug2016analytic} (74\%) but, again, there are significant experimental differences.}

In addition to reporting validation results, we use an online dataset to assess performance for both item classification (Test I) and grasp success prediction (Test II). There was an appreciable (16\%) drop in performance from the offline validation (93\%) to an online test (77\%). This performance drop is partly due to changes which may occur over time and are not captured in training, such as trial-to-trial variation in tension of the springs and cables, or movement of the cameras within the tactile sensors. \textcolor{black}{Despite the small number of online trials per object, the total number of tests (104) demonstrates that the T-MO is capable of robust grasp success prediction and object classification in real time. Moreover these results are more comprehensive than previous work where online tests were not performed.\citep{Flintoff2018Single-GraspSensors, FunabashiObjectRecognition}} We cannot rule out some over-fitting to the training set, although measures were taken to reduce this (Section~\ref{Sec:nn}). A larger dataset would allow exploration of more complicated networks, such as introducing recurrent neural networks for improved use of temporal information rather than the 3D (2 space, 1 time) convolutions used. While more data may improve the accuracy of the online tactile predictions, the priority in this initial study was to establish the T-MO's credibility as a tactile grasping system.

\textcolor{black}{A further tactile analysis was conducted by testing the T-MO with differing numbers of working tactile sensors to demonstrate the benefit of a three-fingered system. When two sensors were active, results were not greatly affected compared to all three sensors. The greatest drop in performance was when the `thumb' was omitted. When using an opposed grasp, the thumb has the highest normal force, which may thus provide more reliable tactile information. Conversely, dropping to just a single sensor resulted in a large performance drop.}

\textcolor{black}{Overall, we sought to maximise performance whilst minimising cost and complexity. Hence, this result demonstrates that three fingers enable high performance even under reduced operating conditions. In contrast, two-finger grippers may be rendered inoperable under the loss of a single tactile sensor, whereas four-plus finger systems add cost and complexity.}

The JeVois camera system used here is small and low cost with machine vision processing modules that are small enough to be embedded within the hand. In principle, onboard processing could be performed via neural networks loaded on the modules using TensorFlow Lite compatibility. This was not utilized in this study, but would give a route to process and react to captured images without the need for a high-performance control PC. Alternatively, a dimensionally-reduced output, such as that from the convolutional layers, could be sent to reduce the processing requirements external to the hand. This would give a semi-autonomous robot hand the ability to interpret its tactile sense.

For an initial exploration of the potential of this onboard processing, we examined whether item classification could be accurately performed on the hand (Section \ref{sec:sensitivity}). The camera processing module are not directly connected to each other, so the tactile images from each sensor must be processed independently. If the mean prediction from all three sensors is used (rather than processing data from all three), the performance drops by just $2\%$ to $91\%$ for item classification (Test I) and $3\%$ to $92\%$ for grasp success prediction (Test II). This shows that deployment of the computation on the JeVois would give a small reduction in performance. Use of the onboard processing would allow the T-MO to be deployed as a complete autonomous system.

\subsection{Hardware Design}
In this study, we use the same 3D-printed skin as other tactile sensors within the TacTip family, with an array of pins spaced approximately 3\,mm apart that were developed originally for a 40\,mm-diameter, domed tactile sensor.\citep{ward2018tactip} This design choice was for consistency in the absence of a good reason to customize those aspects, unlike for example the overall shape of the sensing surface which required customization to fit onto the Model O fingertip. However, it does result in relatively few pins (a $10\times 3$ array) compared with others in the TacTip family.~\cite{ward2018tactip} Since these pins are used as sensing elements, this may limit the tactile sensor performance.

\textcolor{black}{When compared to previous hands with integrated TacTip sensors there are two clear operational benefits provided by the T-MO. Firstly, the T-MO is entirely self-contained with onboard processing allowing computational load on the control PC to be reduced. This is not the case with the M2\citep{ward2016tactile} and Shadow Modular Grasper\citep{Pestell2019AGrasper} hands integrated with TacTips. The tactile integration of the two-fingered GR2 gripper\citep{ward2017gr2} utilised Raspberry Pi computers but these were not integrated into the design, meaning the entire system cannot at present be attached as a single unit to a robotic arm. The second advantage is having three sensors: the T-MO retained high tactile sensing performance when operating with two sensors, but performance dropped significantly for just one sensor; in contrast, any loss of sensing for the GR2 or M2 hands would potentially have much more performance loss.}

There are many design options that may improve tactile sensor performance: the pins could be made smaller, the skin thinner, the pin spacing could be non-uniform, the shape or distribution of the pins could be changed, raised `fingerprints' used\citep{Cramphorn2017AdditionAcuity} \textcolor{black}{or differently shaped sensing area, such as circles.\citep{ward2017gr2} could be tested} It is hard to intuit the effect of these choices and despite the platform being autonomous, it does require human supervision/intervention and so a rigorous examination of multiple designs, while feasible, would be laborious. We anticipate the best design will be task dependent, and so likely different for each tactile benchmark considered here.

\textcolor{black}{For a final comment, we note that the T-MO is able to perform well in comparison with other tactile hands even though it has a much lower cost (costing us less than £1100 to make). We have demonstrated comparable results in tactile perception to previous work that utilizes more expensive robot systems such as the BarrettHand,\citep{dang2014stable} Schunk\citep{LiuObjectRecognition} and iCub\citep{regoli2017controlled} hands. Most of the T-MO's cost is actually its motors, followed by the 3D-printing and camera costs, so it may be possible to reduce these costs still further. A benefit of building the hand ourselves is that we can quickly fix issues such as accidental damage or parts failure, while being 3D printed means the hand can be adapted to various applications and we can fix design problems that only become apparent later in its use.} 

\section{Conclusion}
Overall, we have demonstrated that soft biomimetic tactile sensors embedded within a robot hand -- the Tactile Model O -- are able to accurately distinguish and categorize objects using only tactile information from the finger surface deformation. This ability to classify objects without vision is essential in many scenarios, such as clearing items from a cluttered bin and where a visual snapshot may not be a reliable indicator of the object properties. \textcolor{black}{Although we did find a reduction in grasping performance because of the changes to the shape of the fingers, we believe that this is justified in situations requiring tactile sensing. In particular, the T-MO demonstrated high performance at object recognition and grasp success prediction, with both of these relatively robust to sensor failure. Furthermore, we anticipate that grasping performance could be enhanced by further design changes such as the addition of a fingernail and testing different sensor constructions.}

\textcolor{black}{The T-MO is a low-cost, 3D-printed tactile robot hand with multi-purpose tactile sensing capability and the ability to undertake computation on-board the hand to reduce the high bandwidth requirements associated with images. Additionally, to the authors' knowledge, no other system has demonstrated the ability to perform both grasp stability prediction and object classification in real time and in a single attempt.} We believe this demonstrates that the T-MO is an effective platform for robot hand research, and expect it to open-up a range of applications in autonomous object handling. 


\section*{Acknowledgements}
We thank Kirsty Aquilina, Gareth Griffiths, Raia Hadsell, John Lloyd, Nicholas Pestell, Andrew Stinchcombe and Ben Ward-Cherrier for their help. NL was supported by awards from the Leverhulme Trust on ‘A biomimetic forebrain for robot touch’ (RL-2016-390) and the Engineering and Physical Sciences Research Council (EPSRC) on `Tactile Superresolution Sensing' (EP/M02993X/1). JJ and LC were supported by the EPSRC CDT in Future Autonomous and Robotic Systems (FARSCOPE). AC was supported by Google DeepMind. 

\bibliographystyle{unsrtnat}
{\footnotesize\bibliography{tmo_refs}}

\begin{thebibliography}{49}
\providecommand{\natexlab}[1]{#1}
\providecommand{\url}[1]{\texttt{#1}}
\expandafter\ifx\csname urlstyle\endcsname\relax
  \providecommand{\doi}[1]{doi: #1}\else
  \providecommand{\doi}{doi: \begingroup \urlstyle{rm}\Url}\fi

\bibitem[Johansson and Flanagan(2009)]{johansson2009coding}
Roland~S Johansson and J~Randall Flanagan.
\newblock {Coding and use of tactile signals from the fingertips in object
  manipulation tasks}.
\newblock \emph{Nature Reviews Neuroscience}, 10\penalty0 (5):\penalty0 345,
  2009.

\bibitem[Kappassov et~al.(2015)Kappassov, Corrales, and
  Perdereau]{Kappassov2015TactileReview}
Zhanat Kappassov, Juan-Antonio Corrales, and Véronique Perdereau.
\newblock {Tactile sensing in dexterous robot hands — Review}.
\newblock \emph{Robotics and Autonomous Systems}, 74:\penalty0 195--220, 12
  2015.

\bibitem[Luo et~al.(2017)Luo, Bimbo, Dahiya, and Liu]{luo2017robotic}
Shan Luo, Joao Bimbo, Ravinder Dahiya, and Hongbin Liu.
\newblock Robotic tactile perception of object properties: A review.
\newblock \emph{Mechatronics}, 48:\penalty0 54--67, 2017.

\bibitem[Lee(2000)]{Lee2000TactileChallenges}
Mark Lee.
\newblock {Tactile Sensing: New Directions, New Challenges}.
\newblock \emph{International Journal of Robotics Research}, 19\penalty0 (7),
  2000.

\bibitem[Yousef et~al.(2011)Yousef, Boukallel, and Althoefer]{Yousef2011}
Hanna Yousef, Mehdi Boukallel, and Kaspar Althoefer.
\newblock {Tactile sensing for dexterous in-hand manipulation in robotics - A
  review}.
\newblock \emph{Sensors and Actuators, A: Physical}, 167\penalty0 (2):\penalty0
  171--187, 2011.

\bibitem[Odhner et~al.(2014)Odhner, Jentoft, Claffee, Corson, Tenzer, Ma,
  Buehler, Kohout, Howe, and Dollar]{odhner2014compliant}
Lael~U Odhner, Leif~P Jentoft, Mark~R Claffee, Nicholas Corson, Yaroslav
  Tenzer, Raymond~R Ma, Martin Buehler, Robert Kohout, Robert~D Howe, and
  Aaron~M Dollar.
\newblock {A compliant, underactuated hand for robust manipulation}.
\newblock \emph{The International Journal of Robotics Research}, 33\penalty0
  (5):\penalty0 736--752, 2014.

\bibitem[Ward-Cherrier et~al.(2018)Ward-Cherrier, Pestell, Cramphorn, Winstone,
  Giannaccini, Rossiter, and Lepora]{ward2018tactip}
Benjamin Ward-Cherrier, Nicholas Pestell, Luke Cramphorn, Benjamin Winstone,
  Maria~Elena Giannaccini, Jonathan Rossiter, and Nathan~F Lepora.
\newblock {The TacTip Family: Soft Optical Tactile Sensors with 3D-Printed
  Biomimetic Morphologies}.
\newblock \emph{Soft robotics}, 2018.

\bibitem[Ward-Cherrier et~al.(2016)Ward-Cherrier, Cramphorn, and
  Lepora]{ward2016tactile}
Benjamin Ward-Cherrier, Luke Cramphorn, and Nathan~F Lepora.
\newblock {Tactile Manipulation With a TacThumb Integrated on the Open-Hand M2
  Gripper}.
\newblock \emph{IEEE Robotics and Automation Letters}, 1\penalty0 (1):\penalty0
  169--175, 2016.

\bibitem[Ward-Cherrier et~al.(2017)Ward-Cherrier, Rojas, and
  Lepora]{ward2017gr2}
B~Ward-Cherrier, N~Rojas, and N~F Lepora.
\newblock {Model-Free Precise in-Hand Manipulation with a 3D-Printed Tactile
  Gripper}.
\newblock \emph{IEEE Robotics and Automation Letters}, 2\penalty0 (4):\penalty0
  2056--2063, 10 2017.

\bibitem[Schmitz et~al.(2010)Schmitz, Maggiali, Natale, Bonino, and
  Metta]{Schmitz2010AICub}
A~Schmitz, M~Maggiali, L~Natale, B~Bonino, and G~Metta.
\newblock {A tactile sensor for the fingertips of the humanoid robot iCub}.
\newblock In \emph{2010 IEEE/RSJ International Conference on Intelligent Robots
  and Systems}, pages 2212--2217, 10 2010.

\bibitem[Abd et~al.(2018)Abd, Gonzalez, Colestock, Kent, and
  Engeberg]{Abd2018DirectionHand}
Moaed~A. Abd, Iker~J. Gonzalez, Thomas~C. Colestock, Benjamin~A. Kent, and
  Erik~D. Engeberg.
\newblock {Direction of Slip Detection for Adaptive Grasp Force Control with a
  Dexterous Robotic Hand}.
\newblock In \emph{2018 IEEE/ASME International Conference on Advanced
  Intelligent Mechatronics (AIM)}, pages 21--27, 7 2018.

\bibitem[Koiva et~al.(2013)Koiva, Zenker, Schurmann, Haschke, and
  Ritter]{Koiva2013ASensor}
Risto Koiva, Matthias Zenker, Carsten Schurmann, Robert Haschke, and Helge~J.
  Ritter.
\newblock {A highly sensitive 3D-shaped tactile sensor}.
\newblock In \emph{2013 IEEE/ASME International Conference on Advanced
  Intelligent Mechatronics}, pages 1084--1089, 7 2013.

\bibitem[Jara et~al.(2014)Jara, Pomares, Candelas, and
  Torres]{Jara2014ControlFeedback}
Carlos Jara, Jorge Pomares, Francisco Candelas, and Fernando Torres.
\newblock {Control Framework for Dexterous Manipulation Using Dynamic Visual
  Servoing and Tactile Sensors’ Feedback}.
\newblock \emph{Sensors}, 14\penalty0 (1):\penalty0 1787--1804, 1 2014.

\bibitem[Veiga et~al.(2018)Veiga, Edin, and Peters]{veiga2018hand}
Filipe Veiga, Benoni~B Edin, and Jan Peters.
\newblock {In-Hand Object Stabilization by Independent Finger Control}.
\newblock \emph{arXiv:1806.05031}, 2018.

\bibitem[Iwata and Sugano(2009)]{Iwata2009DesignTWENDY-ONE}
H.~Iwata and S.~Sugano.
\newblock {Design of human symbiotic robot TWENDY-ONE}.
\newblock In \emph{2009 IEEE International Conference on Robotics and
  Automation}, pages 580--586, 5 2009.

\bibitem[Schmitz et~al.(2014)Schmitz, Bansho, Noda, Iwata, Ogata, and
  Sugano]{Schmitz2014TactileDropout}
Alexander Schmitz, Yusuke Bansho, Kuniaki Noda, Hiroyasu Iwata, Tetsuya Ogata,
  and Shigeki Sugano.
\newblock {Tactile object recognition using deep learning and dropout}.
\newblock In \emph{2014 IEEE-RAS International Conference on Humanoid Robots},
  pages 1044--1050, 11 2014.

\bibitem[Ma et~al.(2016)Ma, Spiers, and Dollar]{Ma2016M2Gripper}
Raymond~R Ma, Adam Spiers, and Aaron~M Dollar.
\newblock M2 gripper: Extending the dexterity of a simple, underactuated
  gripper.
\newblock In \emph{Advances in reconfigurable mechanisms and robots II}, pages
  795--805. Springer, 2016.

\bibitem[Dong et~al.(2017)Dong, Yuan, and Adelson]{dongimproved}
S~Dong, W~Yuan, and E~H Adelson.
\newblock {Improved GelSight tactile sensor for measuring geometry and slip}.
\newblock In \emph{2017 IEEE/RSJ Int. Conf. on Intelligent Robots and Systems
  (IROS)}, pages 137--144, 9 2017.

\bibitem[Rojas et~al.(2016)Rojas, Ma, and Dollar]{rojas2016gr2}
Nicolas Rojas, Raymond~R Ma, and Aaron~M Dollar.
\newblock {The GR2 Gripper: An Underactuated Hand for Open-Loop In-Hand Planar
  Manipulation.}
\newblock \emph{IEEE Trans. Robotics}, 32\penalty0 (3):\penalty0 763--770,
  2016.

\bibitem[Donlon et~al.(2018)Donlon, Dong, Liu, Li, Adelson, and
  Rodriguez]{Donlon2018GelSlim:Finger}
Elliott Donlon, Siyuan Dong, Melody Liu, Jianhua Li, Edward Adelson, and
  Alberto Rodriguez.
\newblock {GelSlim: A High-Resolution, Compact, Robust, and Calibrated
  Tactile-sensing Finger}.
\newblock In \emph{2018 IEEE/RSJ Int. Conf. on Intelligent Robots and Systems
  (IROS)}, pages 1927--1934, 10 2018.

\bibitem[Wilson et~al.(2020)Wilson, Wang, Romero, and
  Adelson]{wilson2020design}
Achu Wilson, Shaoxiong Wang, Branden Romero, and Edward Adelson.
\newblock Design of a fully actuated robotic hand with multiple gelsight
  tactile sensors.
\newblock \emph{arXiv preprint arXiv:2002.02474}, 2020.

\bibitem[Spiers et~al.(2016)Spiers, Liarokapis, Calli, and
  Dollar]{Spiers2016Single-GraspSensors}
Adam~J. Spiers, Minas~V. Liarokapis, Berk Calli, and Aaron~M. Dollar.
\newblock {Single-Grasp Object Classification and Feature Extraction with
  Simple Robot Hands and Tactile Sensors}.
\newblock \emph{IEEE Transactions on Haptics}, 9\penalty0 (2):\penalty0
  207--220, 4 2016.

\bibitem[Flintoff et~al.(2018)Flintoff, Johnston, and
  Liarokapis]{Flintoff2018Single-GraspSensors}
Zak Flintoff, Bruno Johnston, and Minas Liarokapis.
\newblock {Single-Grasp, Model-Free Object Classification using a
  Hyper-Adaptive Hand, Google Soli, and Tactile Sensors}.
\newblock In \emph{2018 IEEE/RSJ International Conference on Intelligent Robots
  and Systems (IROS)}, pages 1943--1950, 10 2018.

\bibitem[Regoli et~al.(2017)Regoli, Jamali, Metta, and
  Natale]{regoli2017controlled}
Massimo Regoli, Nawid Jamali, Giorgio Metta, and Lorenzo Natale.
\newblock Controlled tactile exploration and haptic object recognition, 2017.

\bibitem[{Funabashi} et~al.(2018){Funabashi}, {Morikuni}, {Geier}, {Schmitz},
  {Ogasa}, {Torno}, {Somlor}, and {Sugano}]{FunabashiObjectRecognition}
S.~{Funabashi}, S.~{Morikuni}, A.~{Geier}, A.~{Schmitz}, S.~{Ogasa}, T.~P.
  {Torno}, S.~{Somlor}, and S.~{Sugano}.
\newblock Object recognition through active sensing using a multi-fingered
  robot hand with 3d tactile sensors.
\newblock In \emph{2018 IEEE/RSJ International Conference on Intelligent Robots
  and Systems (IROS)}, pages 2589--2595, 2018.

\bibitem[{Liu} et~al.(2016){Liu}, {Guo}, and {Sun}]{LiuObjectRecognition}
H.~{Liu}, D.~{Guo}, and F.~{Sun}.
\newblock Object recognition using tactile measurements: Kernel sparse coding
  methods.
\newblock \emph{IEEE Transactions on Instrumentation and Measurement},
  65\penalty0 (3):\penalty0 656--665, 2016.

\bibitem[{Wan} et~al.(2016){Wan}, {Adams}, and {Howe}]{WanVariabilityGrasping}
Q.~{Wan}, R.~P. {Adams}, and R.~D. {Howe}.
\newblock Variability and predictability in tactile sensing during grasping.
\newblock In \emph{2016 IEEE International Conference on Robotics and
  Automation (ICRA)}, pages 158--164, 2016.

\bibitem[Dang and Allen(2014)]{dang2014stable}
Hao Dang and Peter~K Allen.
\newblock Stable grasping under pose uncertainty using tactile feedback.
\newblock \emph{Autonomous Robots}, 36\penalty0 (4):\penalty0 309--330, 2014.

\bibitem[Krug et~al.(2016)Krug, Lilienthal, Kragic, and
  Bekiroglu]{krug2016analytic}
Robert Krug, Achim~J Lilienthal, Danica Kragic, and Yasemin Bekiroglu.
\newblock Analytic grasp success prediction with tactile feedback.
\newblock In \emph{2016 IEEE International Conference on Robotics and
  Automation (ICRA)}, pages 165--171. IEEE, 2016.

\bibitem[Chorley et~al.(2009)Chorley, Melhuish, Pipe, and
  Rossiter]{chorley2009development}
Craig Chorley, Chris Melhuish, Tony Pipe, and Jonathan Rossiter.
\newblock {Development of a tactile sensor based on biologically inspired edge
  encoding}.
\newblock In \emph{Advanced Robotics, 2009. ICAR 2009. International Conference
  on}, pages 1--6, 2009.

\bibitem[Cramphorn et~al.(2017)Cramphorn, Ward-Cherrier, and
  Lepora]{Cramphorn2017AdditionAcuity}
Luke Cramphorn, Benjamin Ward-Cherrier, and Nathan~F. Lepora.
\newblock {Addition of a Biomimetic Fingerprint on an Artificial Fingertip
  Enhances Tactile Spatial Acuity}.
\newblock \emph{IEEE Robotics and Automation Letters}, 2\penalty0 (3):\penalty0
  1336--1343, 7 2017.

\bibitem[Catalano et~al.(2014)Catalano, Grioli, Farnioli, Serio, Piazza, and
  Bicchi]{Catalano2014AdaptiveSoftHand}
M.G. Catalano, G.~Grioli, E.~Farnioli, A.~Serio, C.~Piazza, and A.~Bicchi.
\newblock {Adaptive synergies for the design and control of the Pisa/IIT
  SoftHand}.
\newblock \emph{The International Journal of Robotics Research}, 33\penalty0
  (5):\penalty0 768--782, 4 2014.

\bibitem[Ajoudani et~al.(2014)Ajoudani, Godfrey, Bianchi, Catalano, Grioli,
  Tsagarakis, and Bicchi]{Ajoudani2014ExploringSoftHand}
Arash Ajoudani, Sasha~B. Godfrey, Matteo Bianchi, Manuel~G. Catalano, Giorgio
  Grioli, Nikos Tsagarakis, and Antonio Bicchi.
\newblock {Exploring Teleimpedance and Tactile Feedback for Intuitive Control
  of the Pisa/IIT SoftHand}.
\newblock \emph{IEEE Transactions on Haptics}, 7\penalty0 (2):\penalty0
  203--215, 4 2014.

\bibitem[Ma and Dollar(2017)]{ma2017yale}
Raymond Ma and Aaron Dollar.
\newblock {Yale OpenHand Project: Optimizing Open-Source Hand Designs for Ease
  of Fabrication and Adoption}.
\newblock \emph{IEEE Robotics {\&} Automation Magazine}, 24\penalty0
  (1):\penalty0 32--40, 2017.

\bibitem[Hackett et~al.(2013)Hackett, Pippine, Watson, Sullivan, and
  Pratt]{Hackett2013AnProgram}
Douglas Hackett, James Pippine, Adam Watson, Charles Sullivan, and Gill Pratt.
\newblock {An Overview of the DARPA Autonomous Robotic Manipulation (ARM)
  Program}.
\newblock \emph{Journal of the Robotics Society of Japan}, 31\penalty0
  (4):\penalty0 326--329, 2013.

\bibitem[Ma et~al.(2013)Ma, Odhner, and Dollar]{Ma2013AHand}
Raymond~R. Ma, Lael~U. Odhner, and Aaron~M. Dollar.
\newblock {A modular, open-source 3D printed underactuated hand}.
\newblock In \emph{2013 IEEE International Conference on Robotics and
  Automation}, pages 2737--2743, 5 2013.

\bibitem[Lepora et~al.(2017)Lepora, Aquilina, and
  Cramphorn]{lepora2017exploratory}
Nathan~F Lepora, Kirsty Aquilina, and Luke Cramphorn.
\newblock {Exploratory tactile servoing with active touch}.
\newblock \emph{IEEE Robotics and Automation Letters}, 2\penalty0 (2):\penalty0
  1156--1163, 2017.

\bibitem[Itti(2019)]{Itti2019JeVoisCamera}
Laurent Itti.
\newblock {JeVois Smart Machine Vision Camera}, 2019.
\newblock URL \url{www.jevois.org}.

\bibitem[Lepora and Ward-Cherrier(2015)]{lepora2015superresolution}
Nathan~F Lepora and Benjamin Ward-Cherrier.
\newblock {Superresolution with an optical tactile sensor}.
\newblock In \emph{Intelligent Robots and Systems (IROS), 2015 IEEE/RSJ
  International Conference on}, pages 2686--2691, 2015.

\bibitem[Lepora et~al.(2019)Lepora, Church, de~Kerckhove, Hadsell, and
  Lloyd]{Lepora2019FromSensor}
Nathan~F. Lepora, Alex Church, Conrad de~Kerckhove, Raia Hadsell, and John
  Lloyd.
\newblock {From Pixels to Percepts: Highly Robust Edge Perception and Contour
  Following Using Deep Learning and an Optical Biomimetic Tactile Sensor}.
\newblock \emph{IEEE Robotics and Automation Letters}, 4\penalty0 (2):\penalty0
  2101--2107, 4 2019.

\bibitem[{Lepora} and {Lloyd}(2020)]{lepora2020optimal}
N.~{Lepora} and J.~{Lloyd}.
\newblock Optimal deep learning for robot touch: Training accurate pose models
  of 3d surfaces and edges.
\newblock \emph{IEEE Robotics Automation Magazine}, 2020.

\bibitem[Calandra et~al.(2017)Calandra, Owens, Upadhyaya, Yuan, Lin, Adelson,
  and Levine]{Calandra2017TheOutcomes}
Roberto Calandra, Andrew Owens, Manu Upadhyaya, Wenzhen Yuan, Justin Lin,
  Edward~H. Adelson, and Sergey Levine.
\newblock {The Feeling of Success: Does Touch Sensing Help Predict Grasp
  Outcomes?}, 10 2017.

\bibitem[Kingma and Ba(2014)]{Kingma2014Adam:Optimization}
Diederik~P. Kingma and Jimmy Ba.
\newblock {Adam: A Method for Stochastic Optimization}.
\newblock 2014.
\newblock URL \url{http://arxiv.org/abs/1412.6980}.

\bibitem[Wang et~al.(2004)Wang, Bovik, Sheikh, and
  Simoncelli]{Wang2004ImageSimilarity}
Z.~Wang, A.C. Bovik, H.R. Sheikh, and E.P. Simoncelli.
\newblock {Image Quality Assessment: From Error Visibility to Structural
  Similarity}.
\newblock \emph{IEEE Transactions on Image Processing}, 13\penalty0
  (4):\penalty0 600--612, 4 2004.

\bibitem[Lee et~al.(2019)Lee, Bollegala, and Luo]{lee2019touching}
Jet-Tsyn Lee, Danushka Bollegala, and Shan Luo.
\newblock "touching to see" and "seeing to feel": Robotic cross-modal
  sensorydata generation for visual-tactile perception, 2019.

\bibitem[Calli et~al.(2015)Calli, Walsman, Singh, Srinivasa, Abbeel, and
  Dollar]{Calli2015BenchmarkingSet}
Berk Calli, Aaron Walsman, Arjun Singh, Siddhartha Srinivasa, Pieter Abbeel,
  and Aaron~M. Dollar.
\newblock {Benchmarking in Manipulation Research: Using the Yale-CMU-Berkeley
  Object and Model Set}.
\newblock \emph{IEEE Robotics {\&} Automation Magazine}, 22\penalty0
  (3):\penalty0 36--52, 9 2015.

\bibitem[Jamone et~al.(2016)Jamone, Bernardino, and
  Santos-Victor]{Jamone2016BenchmarkingSet}
Lorenzo Jamone, Alexandre Bernardino, and Jose Santos-Victor.
\newblock {Benchmarking the Grasping Capabilities of the iCub Hand With the YCB
  Object and Model Set}.
\newblock \emph{IEEE Robotics and Automation Letters}, 1\penalty0 (1):\penalty0
  288--294, 1 2016.

\bibitem[Dollar(2019)]{Dollar2019YCBResults}
A~Dollar.
\newblock {YCB Gripper Assessment Benchmark Results}, 2019.
\newblock URL \url{http://www.ycbbenchmarks.com/world-records/}.

\bibitem[Pestell et~al.(2019)Pestell, Cramphorn, Papadopoulos, and
  Lepora]{Pestell2019AGrasper}
Nicholas Pestell, Luke Cramphorn, Fotios Papadopoulos, and Nathan~F. Lepora.
\newblock {A Sense of Touch for the Shadow Modular Grasper}.
\newblock \emph{IEEE Robotics and Automation Letters}, 4\penalty0 (2):\penalty0
  2220--2226, 4 2019.
\newblock ISSN 23773766.
\newblock \doi{10.1109/LRA.2019.2902434}.

\end{thebibliography}
\end{document}